\documentclass[final,5p,times,twocolumn]{elsarticle}
\usepackage{moreverb,url}
\usepackage[colorlinks=false,bookmarksopen,bookmarksnumbered,citecolor=black,urlcolor=black]{hyperref}
\newcommand\BibTeX{{\rmfamily B\kern-.05em \textsc{i\kern-.025em b}\kern-.08em
T\kern-.1667em\lower.7ex\hbox{E}\kern-.125emX}}
\def\volumeyear{2022}

\usepackage[utf8]{inputenc}
\usepackage{pifont}
\usepackage[caption=false]{subfig} % Perhaps try subcaption
\usepackage{graphicx} % Only used for the example images
\usepackage{amsmath}
\usepackage{bm}
\usepackage{commath}
\usepackage[binary-units=true,separate-uncertainty=true,multi-part-units=single,per-mode=symbol]{siunitx}
\DeclareSIUnit\pixel{P}

\usepackage{tabularx}
\usepackage{multirow}
\usepackage{booktabs}
\newcolumntype{C}{>{\centering\arraybackslash}X} % centered version of "X" type
\usepackage{color, colortbl}
\definecolor{Gray}{gray}{0.9}

\usepackage[dvipsnames]{xcolor}

\usepackage{glossaries} 
\newacronym{HSR}{HSR}{Toyota Human Support Robot}

\begin{document}

\begin{frontmatter}

\title{Motion Planning in Dynamic Environments Using Context-Aware Human Trajectory Prediction}

\author[ori]{Mark Nicholas Finean\corref{equal}}
\ead{mark.finean@hotmail.co.uk}

\author[fer]{Luka Petrovi\'{c}\corref{equal}}
\ead{luka.petrovic@fer.hr}

\author[ori]{Wolfgang Merkt}
\ead{wolfgang@robots.ox.ac.uk}

\author[fer]{Ivan Markovi\'{c}}
\ead{ivan.markovic@fer.hr}

\author[ori]{Ioannis Havoutis}
\ead{ioannis@robots.ox.ac.uk}

\address[ori]{University of Oxford, Oxford Robotics Institute (ORI)\\
Dynamic Robot Systems Group (DRS)\\
23 Banbury Rd, Oxford OX2 6NN, United Kingdom}

\address[fer]{University of Zagreb Faculty of Electrical Engineering and Computing\\
Laboratory for Autonomous Systems and Mobile Robotics (LAMOR)\\
Unska 3, HR-10000, Zagreb, Croatia}

\cortext[equal]{These authors contributed equally.}

\nonumnote{This research was supported by the UK Engineering and Physical Sciences Research Council (EPSRC) through the University of Oxford Centre for Doctoral Training, Autonomous Intelligent Machines and Systems (AIMS) [grant references EP/L015897/1, EP/R512333/1], the European Regional Development Fund (DATACROSS) [grant reference KK.01.1.1.01.0009], and in part by the Human-Machine Collaboration Programme, supported by a gift from Amazon Web Services. \textit{Corresponding author: Mark Finean}
}

%%%%%%%%%%%%%%%%%%%%%%%%%%%%%%%%%%%%%%%%%%%%%%%%%%%%%%%%%%%%%%%%%%%%%%%%%%%%%%%%
\begin{abstract}
Over the years, the separate fields of motion planning, mapping, and human trajectory prediction have advanced considerably. However, the literature is still sparse in providing practical frameworks that enable mobile manipulators to perform whole-body movements and account for the predicted motion of moving obstacles. Previous optimisation-based motion planning approaches that use distance fields have suffered from the high computational cost required to update the environment representation. We demonstrate that GPU-accelerated \textit{predicted composite distance fields} significantly reduce the computation time compared to calculating distance fields from scratch. We integrate this technique with a complete motion planning and perception framework that accounts for the predicted motion of humans in dynamic environments, enabling reactive and pre-emptive motion planning that incorporates predicted motions. To achieve this, we propose and implement a novel human trajectory prediction method that combines intention recognition with trajectory optimisation-based motion planning. We validate our resultant framework on a real-world \acrfull[hyper=false]{HSR} using live RGB-D sensor data from the onboard camera. In addition to providing analysis on a publicly available dataset, we release the Oxford Indoor Human Motion (Oxford-IHM) dataset and demonstrate state-of-the-art performance in human trajectory prediction. The Oxford-IHM dataset is a human trajectory prediction dataset in which people walk between regions of interest in an indoor environment. Both static and robot-mounted RGB-D cameras observe the people while tracked with a motion-capture system. 
\end{abstract}
%%%%%%%%%%%%%%%%%%%%%%%%%%%%%%%%%%%%%%%%%%%%%%%%%%%%%%%%%%%%%%%%%%%%%%%%%%%%%%%%

\begin{keyword}Motion Planning \sep Trajectory Optimisation \sep Trajectory Prediction \sep Dynamic Environments \sep RGB-D Perception
\end{keyword}

\end{frontmatter}

\section{Introduction}

In this work, we focus on the deployment of mobile manipulators in dynamic indoor workspaces, such as a household environment. When robots operate in real-world environments, particularly where humans may co-occupy the workspace, safety is paramount. There is an extensive literature base in the space of `autonomous road vehicles' for understanding and predicting `pedestrian' trajectories to assist motion planning and collision avoidance \citep{Rehder2018, Batkovic2018, Flores2019, Luo2018}. In contrast, less work has focused on accounting for the predicted trajectories of humans when planning whole-body robot motions in indoor environments, motivating the work presented here.

Compared to the static case, dynamic environments pose many additional challenges that need to be addressed for robots to operate safely and efficiently. To perform tasks in the presence of moving obstacles, motion planning calculations must be performed online and \textit{quickly} for a robot to react to environmental changes that would otherwise result in collisions. For reactive behaviour to take place, the robot's perception pipeline must be continuously updated online so that changes can be perceived sufficiently fast for the motion planning pipeline to react in time. In our previous work \citep{FineanIros2021}, we presented an integrated framework to enable such reactive behaviour by using a receding-horizon implementation of GPMP2 \citep{gpmp2} in conjunction with the fast GPU-based perception pipeline within GPU-Voxels \citep{GPUVoxels, GPUVoxelsMobile}. While this work enabled reactive whole-body motion planning in response to a dynamic environment, it lacked understanding of dynamic elements in the environment and any reasoning about how they may move in the future. We further introduced the concept of \textit{predicted composite distance fields} \citep{FineanComposite2020} as a fast method of incorporating the predicted motion of moving obstacles directly into a \textit{distance field} (signed or unsigned) representation of the environment for time-configuration space planning. Using this method, separate environment representations are maintained for each timestep in the planned robot trajectory and moving obstacles are propagated using a motion prior, most commonly a constant-velocity model (CVM). We hypothesised that further improvements could be achieved by firstly leveraging parallelism in the problem and utilising GPUs to perform the `compositing', and secondly by incorporating additional scene insights for more realistic obstacle trajectory predictions.

\begin{figure}[t]
\includegraphics[width=\columnwidth]{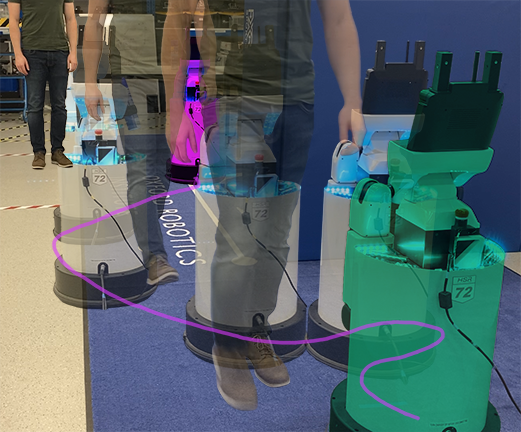}
\caption{A \acrfull[hyper=false]{HSR} is given a whole-body goal to place an object on a table at the other end of the room. A wall obstructs the robot's path and, during execution, a person walks towards a goal located behind the robot. Using our proposed approach and accounting for the predicted trajectory of the person, the robot re-plans to pre-emptively move out of the person's path before continuing towards the goal.}
\label{fig:front_page}
\end{figure}

In this paper, we propose an integrated framework for predictive whole-body motion planning in dynamic environments. For motion planning, we propose the Receding Horizon And Predictive Gaussian Process Motion Planner~2 (RHAP-GPMP2) -- a receding-horizon motion planner that uses composite distance fields to account for the predicted motion of humans in the workspace. We use a state-of-the-art image segmentation method to identify humans and remove dynamic objects from the maintained voxelmap of the static scene. We investigate the task of human motion prediction and propose a planning-based human trajectory prediction method that combines human intention recognition with trajectory optimisation. To explore the problem and aid our analysis of the methods, we further produce and release a dataset for human trajectory prediction that includes robot-perspective RGB-D sensor data. We validate our complete framework in hardware experiments and demonstrate effective collision avoidance across multiple scenarios using a trajectory optimisation-based approach to whole-body motion planning in the presence of moving obstacles; one such scenario is shown in Fig.~\ref{fig:front_page}.

The key contributions of this paper are:
\begin{itemize}
    \item A receding-horizon motion planner that uses composite distance fields to perform time-configuration space motion planning -- Receding Horizon And Predictive Gaussian Process Motion Planner~2 (RHAP-GPMP2).
    \item A novel goal-oriented planning-based human trajectory prediction method that combines human intention recognition with trajectory optimisation.
    \item A comparison of the performance boost provided by GPU-calculated \textit{predicted composite distance fields} with a state-of-the-art algorithm (PBA).
    \item Experimental verification of our integrated framework on an 8-DoF mobile manipulator in 3D dynamic environments using live sensor data.
    \item Release of the Oxford Indoor Human Motion (Oxford-IHM) dataset which comprises human-motion trajectories in an indoor environment, including motion-capture ground truth trajectories, static RGB-D camera images, and RGB-D data captured from the perspective of a moving robot.
    \item An open-source release of our framework which combines human motion prediction with GPU-optimised predicted composite distance fields for trajectory optimisation-based motion planning\footnote{Code available at: \url{https://github.com/ori-drs/integrated-dynamic-motion-planning-framework}}. 
\end{itemize}

\section{Related Work}

In this research, we focus on the development of an integrated framework at the intersection of environment mapping, whole-body motion planning in dynamic environments, and human motion prediction. In the following sections, we review the relevant work across these areas.

\subsection{Perception and Motion Planning in Dynamic Environments}

Although much of the mapping literature has focused on static environments \citep{Newcombe2011, Whelan2012, Whelan2015}, there have been works that consider dynamic environments. 
Static-Fusion \citep{Scona2018} uses geometric clustering to segment and filter out dynamic obstacles from RGB-D images and fuse observations into a dense static reconstruction of the environment. 
PoseFusion \citep{Zhang2020} combines OpenPose \citep{Cao2021} with ElasticFusion \citep{Whelan2015} to reconstruct the static scene while removing humans from the reconstruction. While OpenPose can be used to estimate the positions of body joints of humans within an RGB image, other dynamic scene reconstruction methods consider instance segmentation or optical flow techniques to separate dynamic parts of the scene from the static background. For example, \cite{Xie2021} perform feature-based RGB-D Simultaneous Localisation and Mapping (SLAM) in dynamic environments using Mask R-CNN \citep{He2020} image segmentation and optical flow-based motion detection. While their approach demonstrates state-of-the-art localisation accuracy, they report an average processing time per frame of \SI{0.42}{\second} and ``up to \SI{1.10}{\second} when mask inpainting'' is required. As these numbers show, image segmentation is generally an expensive process and such long computation times may result in behaviours that lack reactivity when deployed on robots in real dynamic environments.
\cite{Zhang2019} similarly use Mask R-CNN as a method of detecting potentially dynamic obstacles with a SLAM framework.

To achieve real-time run-rates for SLAM in dynamic environments, MaskFusion \citep{Runz2019} supplements semantic instance segmentation (Mask R-CNN) with geometric segmentation. ReFusion uses a Truncated Signed Distance Field (TSDF) based mapping approach to build static maps of the environment and filter out dynamic objects by using the residuals ``from the registration and the representation of free space'' \citep{Palazzolo2019}. As with most of the SLAM literature, the aforementioned mapping systems consider the SLAM problem in isolation and do not consider integration with a motion planner. The reverse is also typically true, whereby motion planners in dynamic environments neglect the need for mapping to take place concurrently with live sensing.

\cite{Park2012} propose a parallel optimisation approach to motion re-planning in dynamic environments with ITOMP. To perform collision avoidance, they utilise pre-computed Euclidean Distance Transforms (EDTs) for static obstacle costs and use geometric collision detection to assign dynamic obstacle costs. However, their method was only tested in simulation and neglects consideration of the need to reconstruct the static environment using live sensor data in the presence of dynamic obstacles. 

Voxblox \citep{Oleynikova2017} and FIESTA \citep{FIESTA} propose incremental mapping frameworks and demonstrate them online. While FIESTA uses a kinodynamic path search method \citep{Zhou2019}, Voxblox is integrated with a trajectory optimisation-based motion planner similar to CHOMP \citep{Oleynikova2016replanning}. In both cases, only 3D path planning for aerial vehicles is performed rather than whole-body motion planning as we propose.

GPU-Voxels \citep{GPUVoxels} is a GPU-optimised framework for multiple environment data structures that can be used for collision avoidance. \cite{GPUVoxels} combine their perception pipeline with a D*-Lite motion planner to demonstrate a mobile robot re-planning in response to newly observed objects. However, they do not demonstrate reactive whole-body behaviour in dynamic environments; this is likely due to the `curse of dimensionality' posed by search-based motion planners.  

\cite{GPUVoxelsMobile} built upon the GPU-Voxels framework to explore fast, exact 3D EDT implementations, such as the Parallel Banding Algorithm (PBA) \citep{PBA}. They use this work to perform fast motion planning for aerial robots with potential field and wavefront planners, integrated with a GPU-based perception framework that leverages the parallelism in EDT computations. 

Of particular interest for our research is the additional aspect of accounting for \textit{predicted trajectories}. \cite{Mainprice2013} used learnt human motions to predict the workspace occupancy for usage with STOMP \citep{Kalakrishnan2011} in simulation experiments. \cite{Park2019} proposed I-Planner which similarly uses offline learning of human actions to generate predicted human motions for use in motion planning within the workspace of a 7-DoF robot arm.

To the best of our knowledge, there does not yet exist a fully integrated perception, motion planning, and prediction pipeline that can enable mobile manipulators to predict the trajectories of moving obstacles and subsequently avoid them in whole-body motion planning tasks. We address this in the work presented here.

\subsection{Human Motion Prediction}\label{subsec:humanmotionprediction}
\begin{figure*}[tb]
\centering
\includegraphics[draft=false,width=1.0\textwidth]{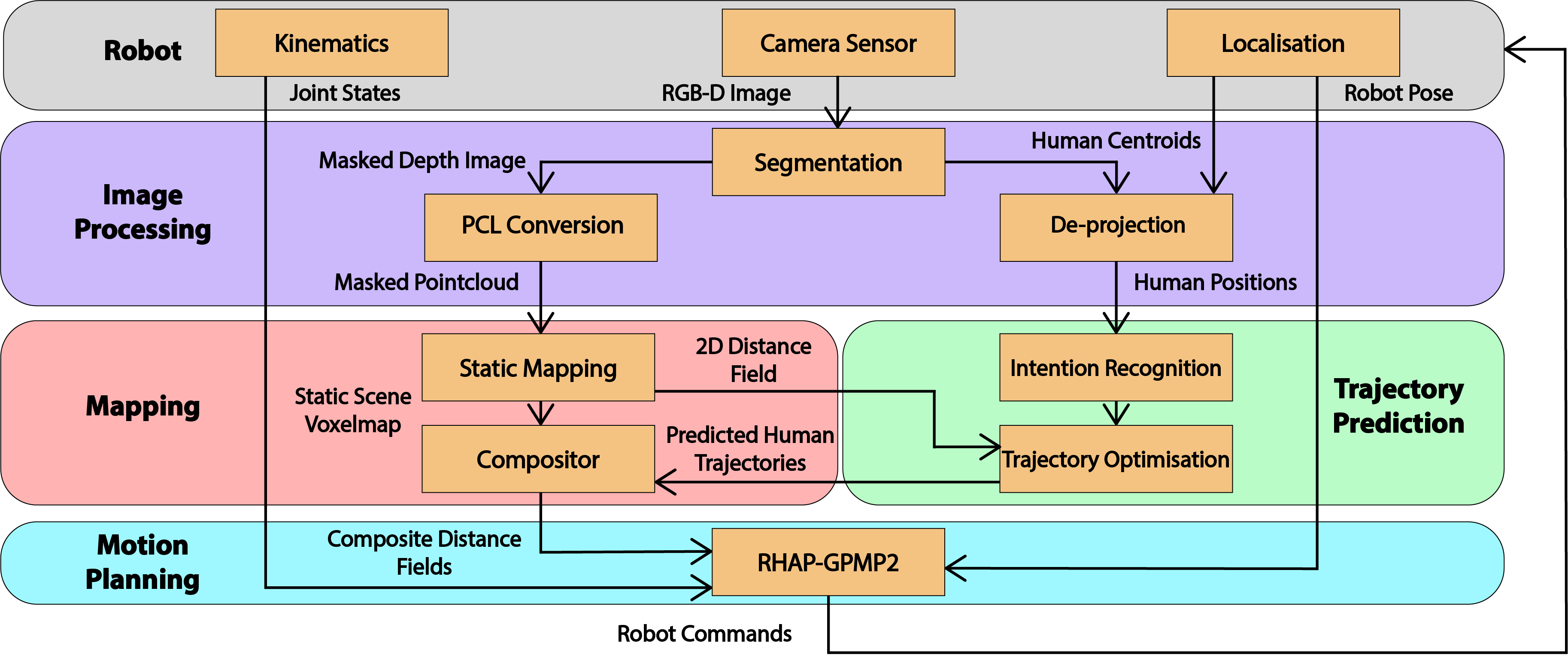}
\caption{Flowchart of our Integrated Framework.}
\label{fig:flowchart}
\end{figure*}
Motion prediction plays an important role in ensuring the safety of robots and autonomous systems; anticipating how objects will move in a scene enables robots to act in a pro-active manner and pre-emptively respond to changes in a dynamic environment to avoid collisions.
For inanimate dynamic objects, such as a rolling ball, we can commonly rely on a purely physics-based model, where simple kinematic models (e.g. constant velocity, constant acceleration) often suffice for enabling collision-free robot operation~\citep{scholler2020constant}.
However, when robots operate in environments alongside humans, safety is of paramount importance and there is a need for more advanced motion prediction to capture the complexity of human behaviour.
This complexity stems from both internal (goal intent, semantics) and external stimuli (environmental priors, actions of surrounding agents) that influence human motion.
The multitude of human motion prediction methods can be categorised by their modelling approach as physics-based, pattern-based and planning-based methods~\citep{rudenko2020human}.

Physics-based methods predict human motion by propagating the current state through an explicit dynamical model~\citep{helbing1995social, treuille2006continuum, pellegrini2009you, zanlungo2011social, karamouzas2009predictive, hermann2015anticipate}.
These methods are typically efficient, interpretable, and work very well for short-term prediction. However, in most cases they do not capture the complexity of the real world, ignoring environmental cues and the possible goals of a person.
Notably, \cite{pellegrini2009you} propose a physics-based method that considers a future destination and the surrounding environment. However, their approach relies on a bird's-eye view and is thus not deployed on a real robot using live sensor data.
\cite{hermann2015anticipate} integrate human trajectory prediction with motion planning and perception by using swept-volume based extrapolation on live RGB-D~data. However, they use the obtained motion prediction only for stopping a robot's movement when potential collisions are detected, rather than performing motion re-planning and adapting the robot's trajectory.

Pattern-based methods attempt to capture human motion behaviour by training function approximators, e.g. neural networks or Gaussian processes, on pre-recorded data~\citep{alahi2016social, ghosh2017learning, bartoli2018context, amirian2019social, yu2020spatio, cao2020long, kratzer2020prediction}.
These models have become dominant in recent years due to their performance for long-term prediction in complex, semantically-rich environments.
However, they require offline learning with large amounts of training data and offer limited transferability to novel environments due to poor generalisation capabilities.
\cite{cao2020long} generate a diverse synthetic dataset to bypass the tedious data collection and propose a learning framework that exploits scene context to improve generalisation; however, their method is computationally demanding and is not deployed on a real robot.
\cite{kratzer2020prediction} utilise a recurrent neural network for encoding short-term dynamics and account for environmental constraints with trajectory optimisation; this method disregards the existence of multiple possible goals and is not demonstrated in real environments. 

Planning-based approaches assume that a person is moving through an environment towards an existing goal while avoiding obstacles~\citep{ziebart2009planning, vasquez2016novel, rosmann2017online, kitani2012activity, best2015bayesian, rudenko2018joint, zhi2021probabilistic}.
These approaches offer a good balance between long-term prediction performance and capacity for generalisation, but in most cases they require an explicitly defined static map of the environment with the possible goal locations provided, making them difficult to apply on a real robot in an unknown environment.
\cite{best2015bayesian} present a Bayesian framework for intention estimation and use probabilistic roadmaps to obtain trajectory predictions.
However, as a consequence of using a sampling-based planning method that does not account for smoothness, the predicted trajectories exhibit rapid changes of direction that are uncharacteristic of humans.
\cite{ziebart2009planning} predict goal-directed behaviours of pedestrians by solving a soft-maximum Markov Decision Process (MDP) with maximum entropy inverse reinforcement learning~\citep{ziebart2008maximum}.
They demonstrate real-time robot operation that accounts for human motion prediction but their method has limited generalisation capabilities since it relies on learning human reward functions from observed data.

To ensure safe robot operation in indoor environments, we utilise context-specific information and devise a novel trajectory optimisation-based method for human motion prediction that respects the underlying dynamical model and environmental cues. Our proposed method, as detailed in Secs. \ref{sec:int_rec} and \ref{sec:humantrajoptimisation}, offers a hybrid approach that can be deployed in unknown environments without any prior knowledge but can also incorporate information acquired both offline and online by learning possible human goals from observed data. 

\section{Proposed Framework Overview}\label{sec:frameworkoverview}
\begin{figure*}[tb]
          \subfloat[\label{fig:markers_a}]{
          \includegraphics[width=0.66\columnwidth]{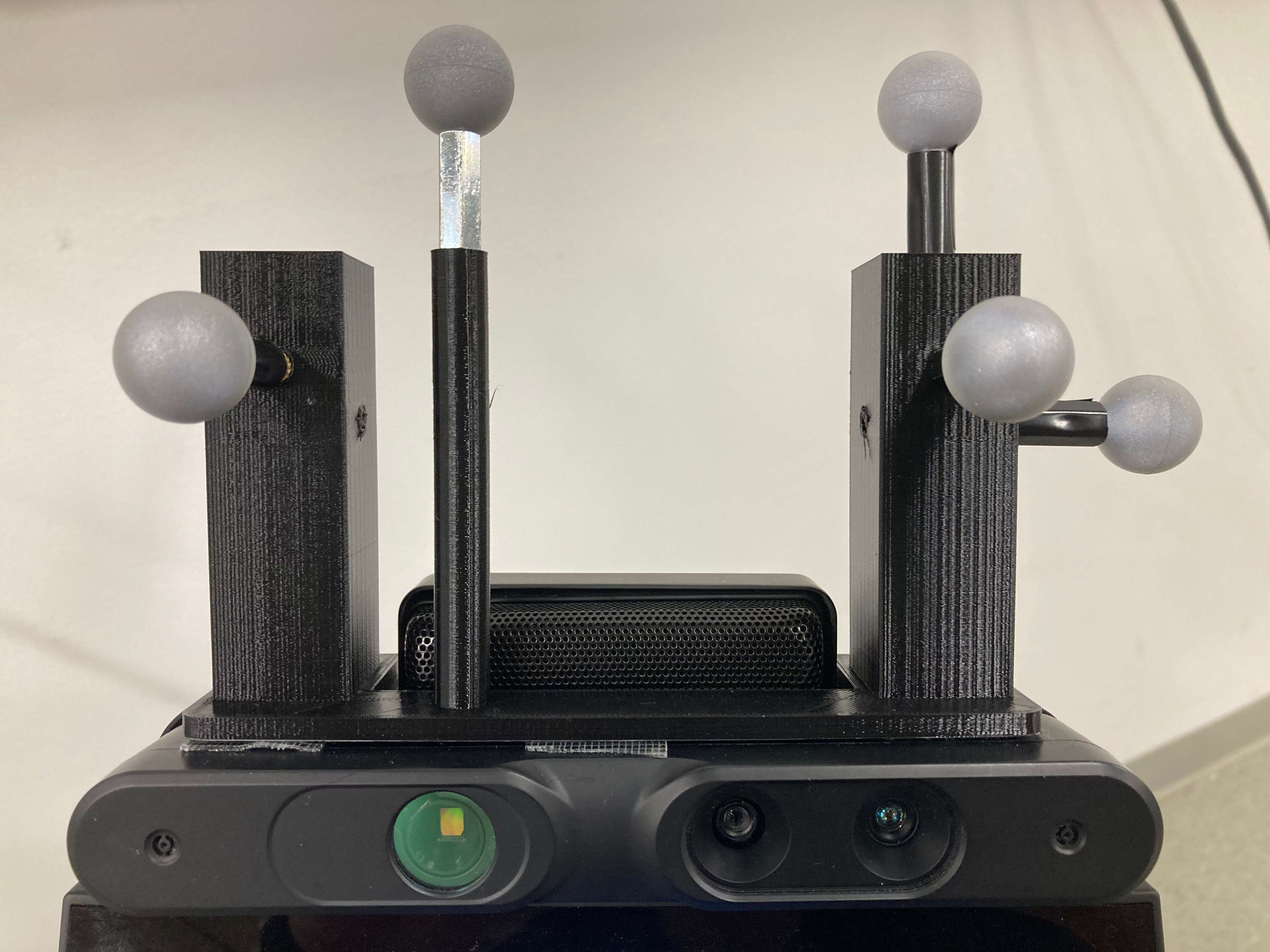}}   
          \hfill
          \subfloat[\label{fig:markers_b}]{
          \includegraphics[width=0.66\columnwidth]{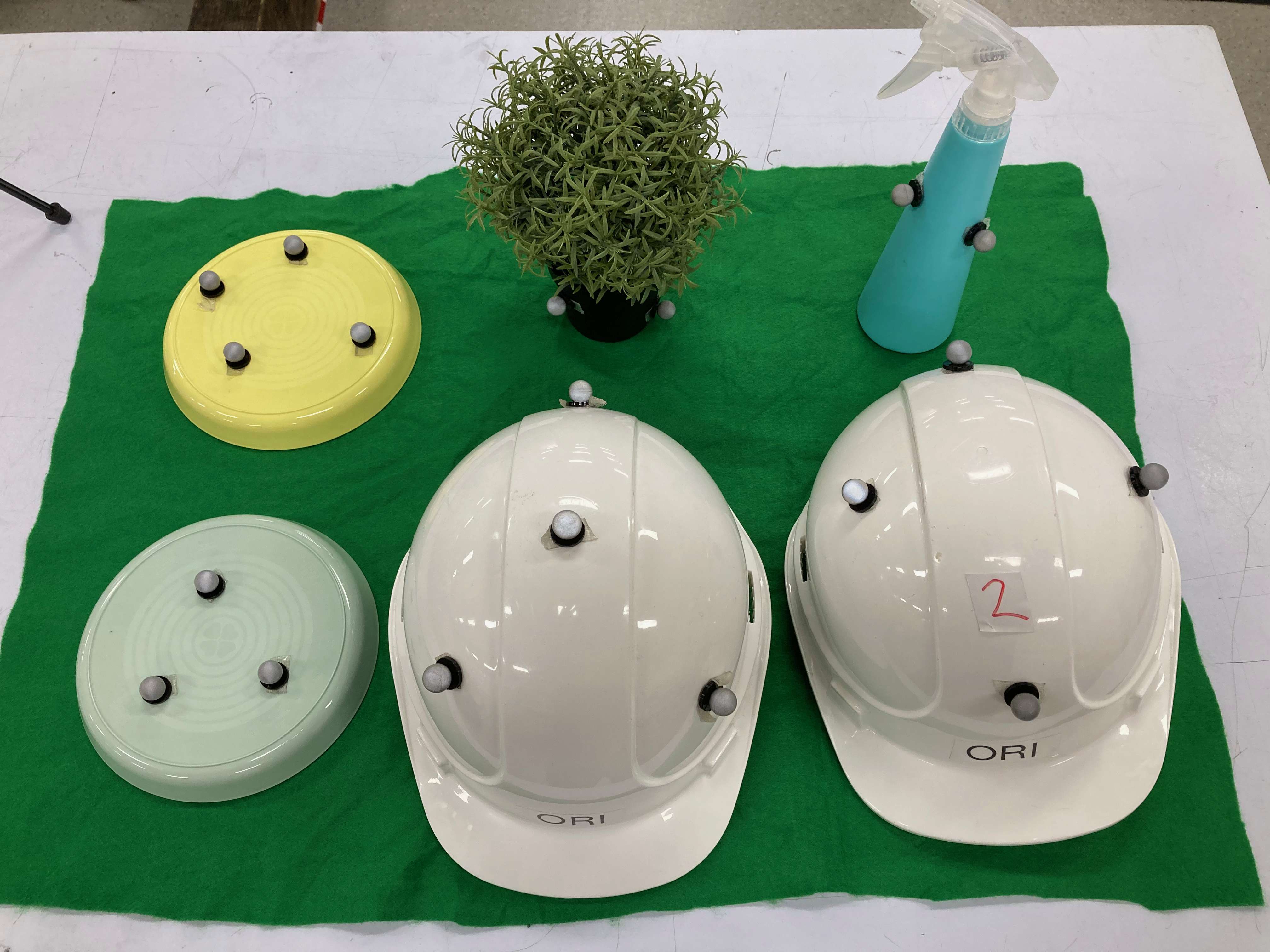}}   
          \hfill
          \subfloat[\label{fig:markers_c}]{
          \includegraphics[width=0.66\columnwidth]{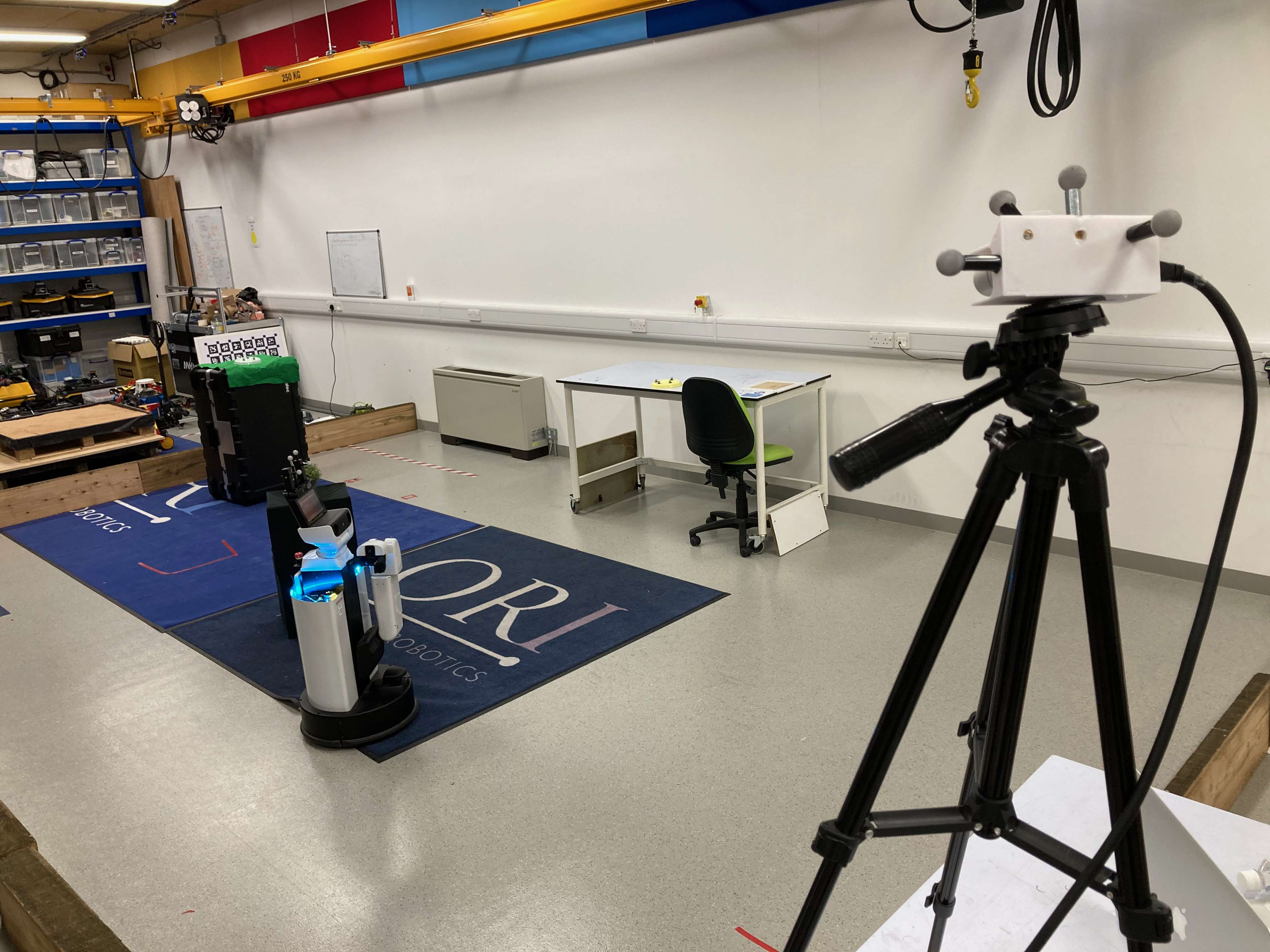}}  
\caption{Photos of the marker arrangements used for motion capture tracking of the robot, environment, and humans. Left: A custom 3D-printed frame for mounting markers on the head of an HSR robot. Middle: Multiple objects were tracked and placed in the environment to provide human `goal' markers. Additional markers were placed at the two entrance/exit locations of the arena. Right: An example setup of the static RGB-D camera within the scene. A custom camera housing was made to attach and calibrate the Vicon markers.}
\label{fig:vicon_objects}
\end{figure*}
In this section, we outline the modular nature of our proposed framework, as illustrated in Fig.~\ref{fig:flowchart}, before examining each module in greater detail. 

The first part of our proposed framework processes RGB~-D images from the camera sensor to extract the information required by other modules.
Using a neural network-based instance segmentation method (described in Sec.~\ref{sec:image_segmentation}) on the RGB images, we generate masks of objects that are detected in the scene. While our method can be trivially adapted to other identified obstacles, in this research, we focus solely on utilising the masks of detected humans. If masks are produced above a user-specified threshold score, $\alpha_{m a s k}$, we apply the masks to their corresponding depth images; we do this to first extract the estimated position of the humans in the scene, and secondly as a method of removing dynamic obstacles from the scene prior to converting depth images to point clouds. The filtered pointclouds are used to provide live updates to the maintained voxelmap of the static scene, while the extracted positions of people in the scene are passed to the trajectory prediction module.

Using the estimated human positions provided by the image processing module, we perform human trajectory prediction.
As described in Sec.~\ref{subsec:humanmotionprediction}, being able to account for the predicted motion of humans is important for safe robot task execution. While the future trajectory of an inanimate object can often be predicted using constant-velocity or constant-acceleration models, human behaviour is more complex and requires a different modelling approach. We propose a hybrid trajectory prediction module (detailed in Sec.~\ref{sec:human_trajectory_prediction}) that uses a lightweight planning-based approach to perform prediction while retaining the ability to incorporate `learnt' or \textit{prior} information. The proposed trajectory prediction method can rely solely on information obtained from live sensor data, requiring no prior training or initialisation.
As shown in Fig.~\ref{fig:flowchart}, our trajectory prediction module can be divided into \textit{Intention Recognition} (Sec.~\ref{sec:int_rec}) and \textit{Human Trajectory Optimisation} (Sec.~\ref{sec:humantrajoptimisation}).

After predicting trajectories for humans in the scene, we forward these predictions to the mapping module so that the predicted positions of people in the scene can be composited into distance fields that are maintained for each timestep in our proposed motion planning algorithm, Receding Horizon And Predictive Gaussian Process Motion Planner~2 (RHAP-GPMP2). With the composite distance fields being continuously updated from the latest observations and trajectory prediction information, our motion planning algorithm is able to re-plan and enable reactive, pre-emptive robot behaviours to avoid moving people in the workspace. 

\section{Human Motion Dataset}\label{sec:dataset}
\begin{figure}[t]
\includegraphics[width=\columnwidth]{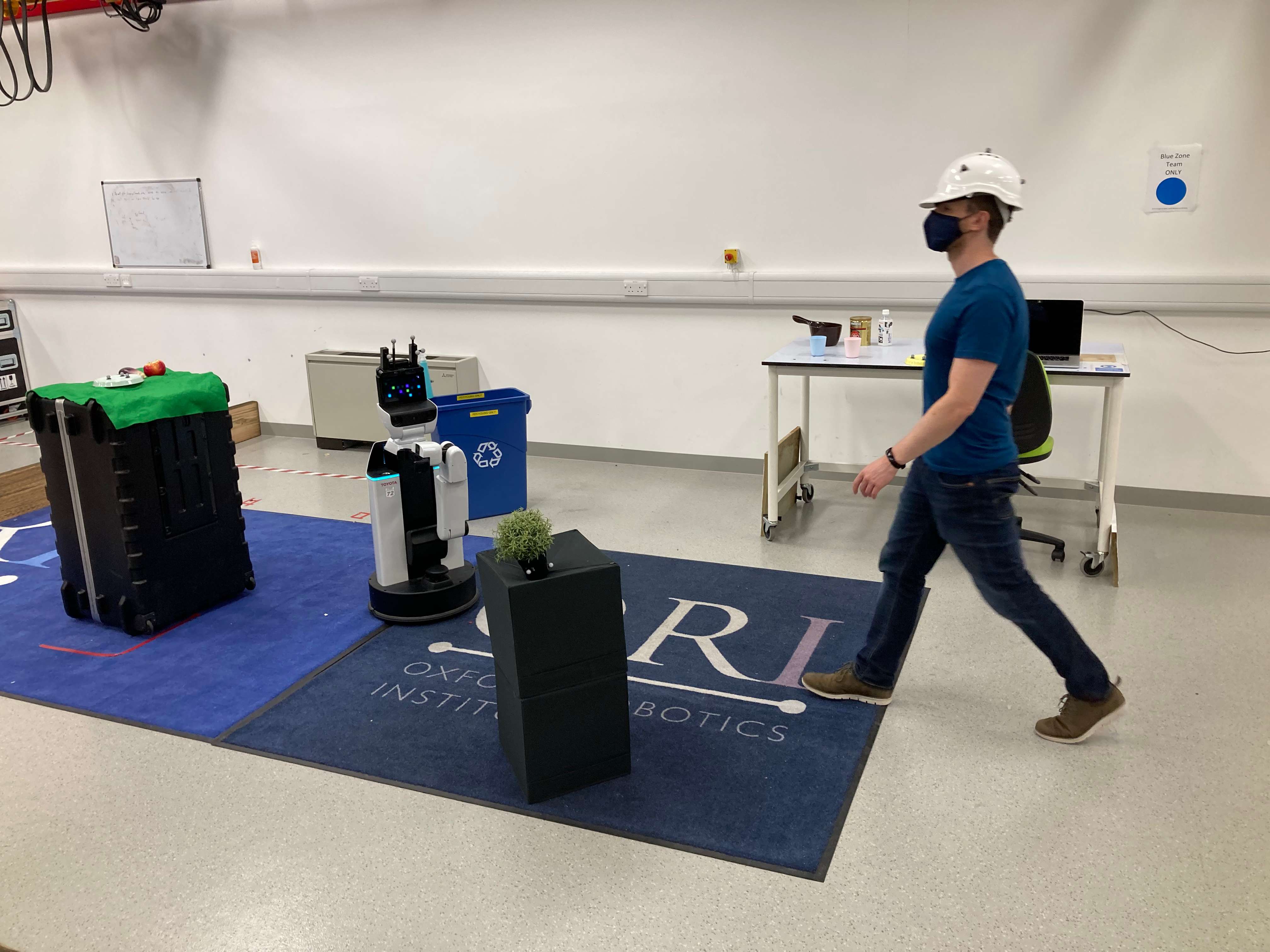}
\caption{An example map configuration used in the Oxford-IHM dataset. As the Vicon-tracked person walks between goals in the scene, the HSR robot is manually controlled by a human operator to move around the scene and maintain vision of the person.}
\label{fig:dataset_scene}
\end{figure}
For this research, we are concerned with robots that operate in indoor environments co-occupied by humans. In order to evaluate any proposed method for human motion prediction, we require an appropriate dataset that, in line with our ambitions for robot autonomy, contains sensor data from the robot's perspective.

While there are a number of publicly available RGB-D datasets, the majority are aimed at applications of SLAM \citep{Sturm2012, Handa2014}, object detection \citep{Janoch2011, Lai2011}, or human activity recognition \citep{Sung2012, Zhang2011}. \cite{Sturm2012} present an RGB-D dataset where sensor data is collected from the robot's perspective for the purpose of evaluating SLAM systems. However, the dataset lacks the presence of moving humans and their ground truth trajectories for which we can try to predict.

\cite{Munaro2014} released the most relevant dataset for our purposes, the Kinect Tracking Precision (KTP) Dataset. The dataset comprises RGB-D images recorded from a robot's perspective in a scene where humans are moving around while the robot performs locomotion. Totalling only four minutes of video recording, we do not believe that the KTP dataset captures an accurate representation of human motion over a wide enough range of behaviours for prediction purposes. In their recordings, humans move either in a linear or random manner. While this may be how human motion appears sometimes, we believe that humans usually have \textit{intent}, i.e. a destination and task in mind. For example, people will often travel to their desks, a bookcase, or exit through a door in an office environment. Before people enter a room, we could intuitively generate a prior map of where they will go and refine our belief as their trajectory progresses. 

The concept of \textit{intent} motivates a dataset in which humans are performing tasks relevant to the environment. The TH\"OR dataset \citep{thorDataset2019} provides human motion trajectories and broad goal locations within an indoor experiment; however, it uses 3D LiDAR scans from a stationary sensor rather than RGB-D data from a robot perspective. 
To address the need for a robot perspective RGB-D dataset with task-based human motion trajectories, we propose and release the Oxford Indoor Human Motion (Oxford-IHM) Dataset\footnote{Dataset available at \url{https://ori-drs.github.io/oxford-indoor-human-motion-dataset}}. 
We summarise a selection of the most relevant publicly available datasets that include human trajectories, alongside our own, in Table \ref{table:trajectory_dataset_comparison}.

\begin{table*}[tb]
\small
  \renewcommand{\arraystretch}{1.25}% Spread rows out...
\begin{tabularx}{\textwidth}{| C | c | c | C | c | c | c | c |}
\hline
Dataset & Environment & Duration & External Sensors & Robot Perspective & Motion Capture & Goals & Map \\ 
\hline
KTP \citep{Munaro2014} & Empty Room & 4m 40s & \ding{56} & RGB-D & \ding{51} & \ding{56} & \ding{56} \\
ATC \citep{Brscic2013} & Shopping Centre & 41 days & RGB-D & \ding{56} & \ding{56} & \ding{51}  & \ding{56} \\
TH{\"O}R \citep{thorDataset2019} & Laboratory & 60 mins & 3D LiDAR, RGB, Eye Tracking & \ding{56}  & \ding{51} & \ding{51}  & \ding{51} \\
L-CAS \citep{yan2017online} & Office & 49 mins & \ding{56} & 3D LiDAR & \ding{56} & \ding{56} & \ding {56} \\
% GTA-IM \citep{GTAIM} & Household (Simulated) & x mins & - & -  & - & - -\\
% Freiburg \citep{Sturm2012} & Indoor & Office + Warehouse & - mins & RGB-D & RGB-D & Yes & No \\
% BIWI Walking Pedestrians \citep{Pellegrini2009} & Outdoor & - & - & - & -  & - \\
% MoVI \citep{Ghorbani2021} & Indoor & Laboratory Room & 9hrs & RGB & Handheld RGB  & No & No \\
MoGaze \citep{Kratzer2021} & Laboratory & 180 mins & Eye Tracking & \ding {56} & \ding {51} & \ding {51} & \ding {51} \\
\hline
\bf{Oxford-IHM} & Laboratory/Office & 60 mins & RGB-D & RGB-D & \ding{51}  & \ding{51}  & \ding{51} \\
\hline
\end{tabularx}
\caption{A Comparison of Publicly Available Indoor Datasets with Ground Truth Human Trajectories}
\label{table:trajectory_dataset_comparison}
\end{table*}

\subsection{Data Acquisition}

Our dataset was recorded in a large indoor laboratory within which we constructed an arena similar to an office environment. Under a Vicon motion capture setup, we created the arena of interest, measuring $\SI{7.1}{\metre} \times \SI{4.2}{\metre}$, with perimeter walls and two entrance/exit locations. Within the arena, multiple large objects, such as a desk, were arranged in multiple configurations to act both as potential goals and static obstacles. As a tracked human walks between goals in the arena, we recorded RGB-D images from both a static Intel Realsense D435 camera (Fig.~\ref{fig:markers_c}) and a Toyota Human Support Robot's head-mounted ASUS Xtion Pro Live camera. Additionally, we recorded the robot's \texttt{tf} data which details the robot's 3D pose and joint transformations over time.

While the robot supports an Ethernet connection for data transfer, to avoid trailing cables and maintain recording bandwidth, we opted to control the robot wirelessly and record robot data locally. We used \textit{chrony} time-synchronisation between the robot's onboard clock and an external laptop (Intel Core i7-10875H CPU, \SI{32}{\giga\byte} \SI{2666}{\mega\hertz} RAM, and an NVIDIA GeForce RTX 2070 SUPER GPU). The external laptop was used to record Vicon marker data and RGB-D image data from the static Realsense camera.

During recording, the robot was remotely controlled and navigated around the arena, generally in such a manner that it maintained vision of the tracked person. The overhead motion capture setup was used to record the ground truth locations of goals, entrance/exit locations, the robot, and the person in the scene. The motion capture arrangement consisted of 18 Vicon Vero 2.2 cameras (\SI{2.2}{\mega\pixel} with \SI{850}{\nano\metre} IR emitters). We calibrated the cameras using a Vicon Active Wand v2 to achieve a sub-millimetre average residual tracking error. For accurate tracking, unique IR marker configurations were affixed to each trackable object. For person tracking, reflective markers were attached to helmets and calibrated to align each helmet's orientation with a person's gaze direction. In the cases of the robot and the external camera, custom mounts were 3D printed (Fig.~\ref{fig:vicon_objects}) and used to calibrate each object's tracked pose with its respective internal camera frame.

Our dataset consists of $\approx 60$ minutes of rosbag data split approximately equally across four different map configurations, each with three runs. For each map configuration, we used the ROS \textit{hector\_mapping} package \citep{hectormapping} and the robot's onboard Hokuyo UST-20LX laser range sensor to produce a 2D map of the arena. To provide additional variation, our dataset uses two people with the map configurations split equally between them.

\section{Image Processing}
\label{sec:image_segmentation}

As discussed in Sec.~\ref{subsec:humanmotionprediction}, numerous approaches have been explored in previous efforts to perform dense environment mapping in dynamic environments \citep{Scona2018, Zhang2020, Cao2021, Palazzolo2019}, with many employing image segmentation techniques \citep{Xie2021, Runz2019, Zhang2019}.

\subsection{Image Segmentation}
\begin{table*}[tb]
\small
\begin{tabularx}{\linewidth}{|c|C|C|C|C|C|C|C|C|}\cline{2-9}
\multicolumn{1}{c|}{}&\multicolumn{2}{c|}{MOTP (m)}&\multicolumn{2}{c|}{MOTA (\%)}&\multicolumn{2}{c|}{FP (\%)}&\multicolumn{2}{c|}{FN (\%)} \\
\hline
Task & Ours & KTP & Ours & KTP & Ours & KTP & Ours & KTP \\ 
\hline
Back and Forth & $\bm{0.176}$   & $0.196$       & $\bm{91.7}$    & $88.97$ & $5.9$ & $\bm{2.4}$ & $\bm{1.1}$ & $8.5$ \\
Random Walk    & $0.171$        & $0.171$       & $\bm{76.0}$    & $70.93$ & $\bm{2.3}$ & $9.8$ & $\bm{8.6}$ & $18.9$ \\ 
Side-by-Side   & $0.151$        & $\bm{0.146}$  & $85.9$    & $\bm{87.22}$ & $\bm{0.7}$ & $1.2$ & $\bm{6.7}$ & $11.6$ \\ 
Running        & $\bm{0.136}$   & $0.143$       & $91.6$    & $\bm{94.57}$ & $\bm{0.0}$ & $1.1$ & $\bm{4.2}$ & $4.4$ \\ 
Group          & $0.198$        & $\bm{0.181}$  & $\bm{59.4}$    & $47.91$ & $\bm{1.7}$ & $9.1$ & $\bm{15.8}$ & $42.53$ \\ 
\hline
\end{tabularx}
\caption{Image Processing and Human Position Estimation Benchmarking. For comparison, the KTP performance metrics are replicated from \citep{Munaro2014}.}
\label{table:image_process_bench}
\end{table*}
A commonly used state-of-the-art method of image segmentation is Mask R-CNN~\citep{He2020}. Mask R-CNN is an extension of Faster R-CNN \citep{Ren2017} that predicts the mask of an object in parallel with bounding box recognition. Image segmentation is typically an expensive operation to perform -- \cite{He2020} reported a frame rate of \SI{5}{\hertz} on an NVIDIA Tesla M40 GPU. For our purposes of using online environment reconstructions for motion planning in dynamic environments, we require minimal latency between making sensor observations and them being reflected in a motion planner's collision-checking ability. As such, in this work we build on recent advances in image segmentation performance.

While numerous works have sought to improve Mask R-CNN, few works focus on improving the speed of the instance segmentation \citep{Lee2020}. \cite{Lee2020} introduced CenterMask and CenterMask-Lite, anchor-free one-stage instance segmentation methods that outperform the current state-of-the-art -- the authors report that CenterMask-Lite with a VoVNetV2-39 backbone achieves a frame rate of \SI{35}{\hertz} on an NVIDIA Titan Xp GPU.
We performed a local benchmarking of the latest release, CenterMask2-Lite, against Mask R-CNN on an RGB-D camera stream and found the lightweight CenterMask2-Lite to run $3.2\times$ faster than Mask R-CNN -- \SI{13.4}{\hertz} compared with \SI{4.2}{\hertz}. This test was performed using: NVIDIA RTX 2060 GPU, 8-core Intel Core i7-9700 CPU @ \SI{4.50}{\giga\hertz} and \SI{2133}{\mega\hertz} DDR4 RAM.

Due to the importance of fast perception and motion planning when operating in a dynamic environment, we elected to exploit the enhanced performance provided by CenterMask2-Lite for image segmentation in our perception pipeline. 

\subsection{Human Position Estimation}
As described previously, and illustrated in Fig.~\ref{fig:flowchart}, we perform image segmentation on a stream of RGB images and use the results for both maintaining the static representation of the environment and for estimating the position of humans in the scene. For each RGB frame that contains masks labelled as a \textit{person} with a score above the specified threshold, $\alpha_{m a s k}$, we apply the masks to the corresponding time-synchronised depth image. In this work, we found $\alpha_{m a s k} = 0.7$ to perform well in consistently masking people even when partially obstructed by obstacles.
Using the masked depth image, we extract a depth to associate with the person. In this work, we use the median depth and pixel position of a person's mask to calculate the person's 3D position in the workspace -- this position is passed onto the tracking and prediction module.

\subsection{Object Masking and Pointcloud Conversion}
To filter out the dynamic element of the scene, we apply valid \textit{person} masks to their corresponding depth images. The filtered depth images are converted to pointclouds for integration into the maintained voxelmap of the static environment. We found that it is beneficial to apply a \texttt{dilation} to the \textit{person} masks before converting to pointclouds. By enlarging the segmentation masks, we reduce leakage from the dynamic masks into the static voxelmap. We use OpenCV to perform four iterations of dilation with a $5 \times 5$ kernel.

\subsection{Evaluation}
To validate our image processing pipeline and demonstrate its effectiveness, we evaluate our method on the KTP Dataset \citep{Munaro2014}, using the same metrics as used in their evaluation \citep{Bernardin2008}. Due to the ground truth positions in this dataset corresponding to a person's tracked head location, \cite{Munaro2014} use the ``centroid of the cluster points belonging to the head of the person'' and add a \SI{10}{\centi\metre} offset in the viewpoint direction. To provide a similar comparison, rather than using the entire \textit{person} mask for position extraction, we use the top \SI{10}{\%} of the mask to correspond with the head. We similarly add an offset in the viewpoint direction and found \SI{12.5}{\centi\metre} to give the best results.

The Multiple Object Tracking Precision (MOTP) metric indicates the ability of a `tracker' to estimate object positions accurately. The Multiple Object Tracking Accuracy (MOTA) metric indicates the reliability of a tracker to identify objects in an image frame. While we explore instance segmentation in this work, we do not perform `tracking' between frames -- as such, we assume no incorrect object associations in the calculation of the MOTA metric. Additional recorded metrics of interest are the False Positive (FP) and False Negative (FN) rates. Our benchmarking results are presented in Table~\ref{table:image_process_bench}. 

Importantly for this research, we produce similar MOTP values, indicating that our method achieves a similarly competitive accuracy in estimating the position of people in a scene while additionally providing masks for use in the environment reconstruction. In the following section, we discuss how the extracted positions of people feed into our trajectory prediction pipeline.

\section{Mapping -  Predicted Composite Distance Fields}\label{sec:sdfs}
\begin{figure}[htb]
          \centering
          \subfloat[\label{fig:composite_a}]{
          \includegraphics[width=0.85\columnwidth]{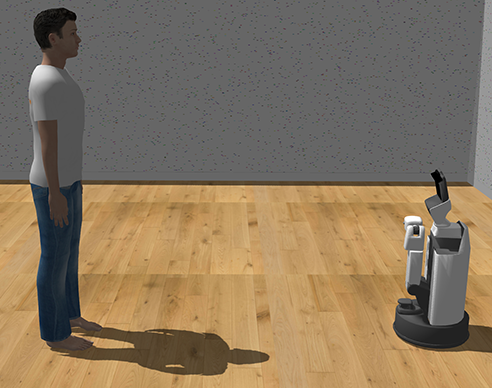}}   
          \vfill
          \subfloat[\label{fig:composite_b}]{
          \includegraphics[width=0.85\columnwidth]{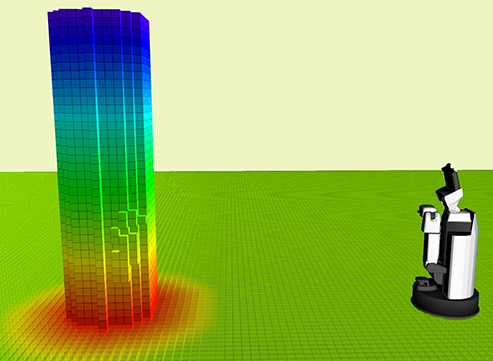}}   
\caption{Top: An example scene in simulation (Gazebo) in which a person is detected by the robot's onboard RGB-D camera. Bottom: The resultant 3D occupancy grid after thresholding the composite distance field for the scene; the distance field of a cylinder is has been composited onto the detected position of the person.}
\label{fig:composite_sdf}
\end{figure}
\begin{figure*}[htb]
          \centering
          \subfloat[Including Device-Host Transfer Time\label{fig:composite_times_inc_transfer}]{
          \includegraphics[width=1.0\columnwidth]{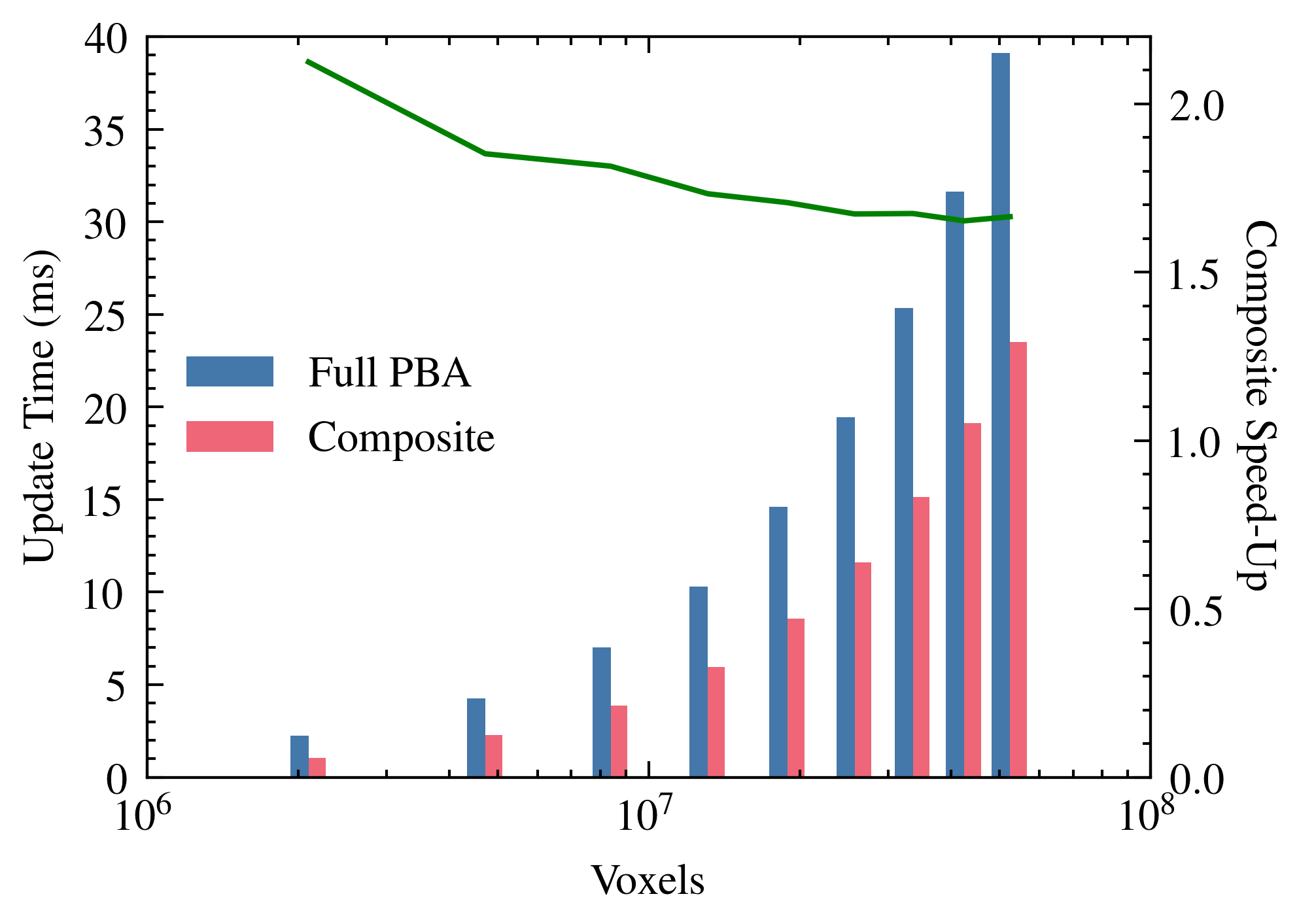}}
          \hfill
          \subfloat[Excluding Device-Host Transfer Time\label{fig:composite_times_exc_transfer}]{
          \includegraphics[width=1.0\columnwidth]{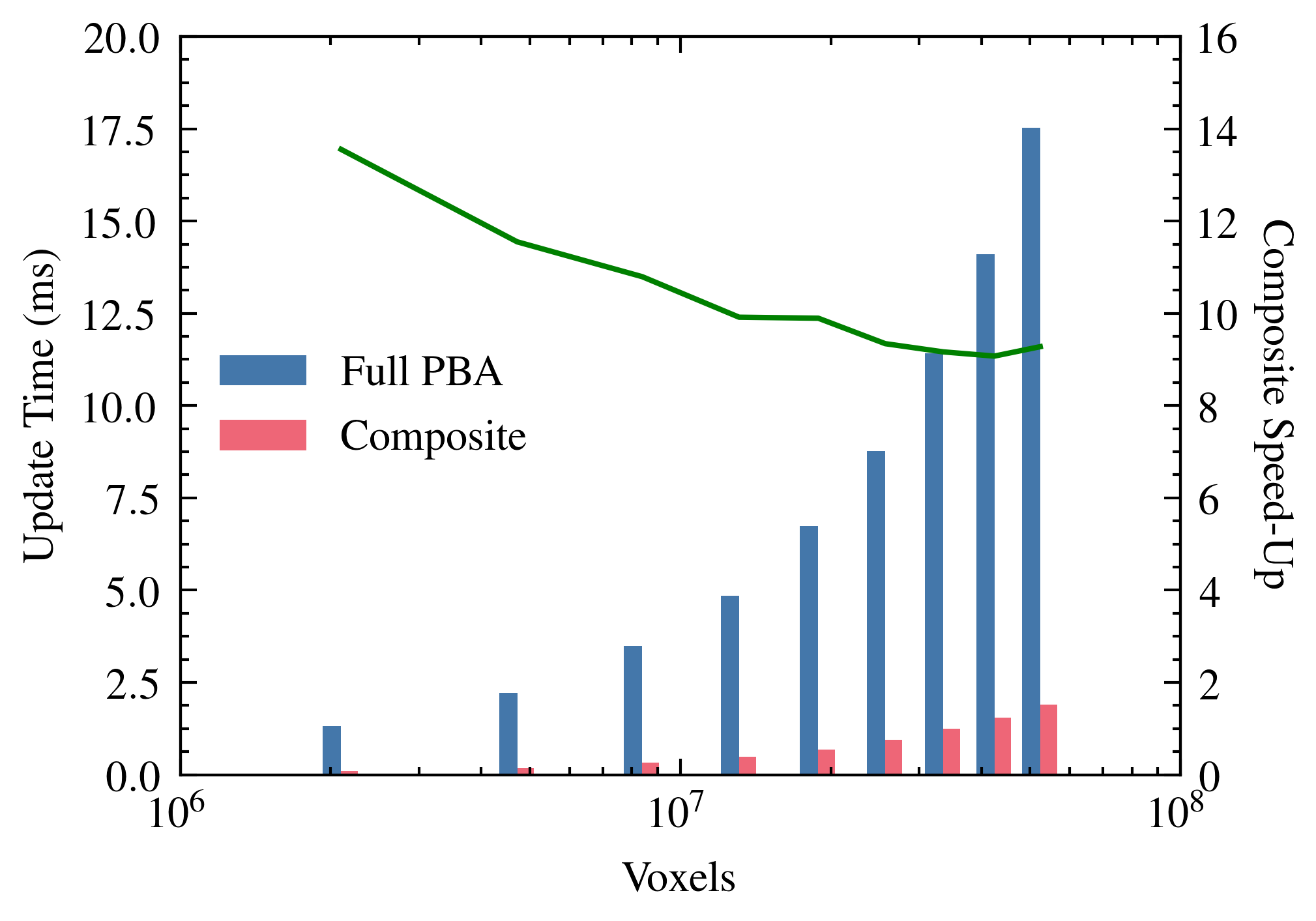}}
\caption{A comparison of the compute times for composite distance fields and full PBA calculations across a range of voxelmap sizes. The plotted line (green) corresponds to the ratio between the bars, i.e. the resultant speed-up of using composite distance fields. While we see an order of magnitude speed-up in the underlying distance field generation, we find that the device-host transfer time dominates the update time, reducing our overall speed-up from $\sim 9.1-13.6\times$ to $\sim1.7-2.1\times$.}
\label{fig:composite_sdf_benchmarking}
\end{figure*}
In our preliminary work on \textit{predicted composite distance fields} \citep{FineanComposite2020}, we suggested that further computational gains could be achieved by performing \textit{compositions} within a GPU-leveraged framework since the core operation of the method is the \textit{min} operation and thus highly parallelisable. In this work, we explore the gains achievable with such an implementation.

Two components are required to generate composite distance fields. Firstly, one needs to maintain a distance field for the environment. Depending on the problem, this may be a static distance field that is computed once at the start of the experiment or continuously updated and maintained as in our framework. Secondly, we require a distance field associated with each (moving) object that is to be \textit{composited} onto the environment distance field. These distance fields can similarly be continuously updated to represent a live model of the obstacles being tracked. In this work, we are interested in human collision avoidance and so the fine voxelised detail of a human is not necessary; instead, we represent humans with similarly sized cylinder shape primitives. The use of primitive shapes is beneficial since we do not need to be concerned with monitoring the shape of the humans and maintaining a live model; rather, we only need to compute the distance field of the primitive shape once and subsequently track the human positions. However, we note that using shape primitives is a choice in this work rather than a limitation -- the distance field could equally be continuously updated to represent an accurate model of a dynamic obstacle as it is observed. There is a large literature base on the dense reconstruction of deformable objects \cite{Dou2015, Yu2015}. 

Given the predicted positions for all dynamic obstacles in the scene for a given time, we can perform a composition of the aforementioned distance fields. This is achieved with a parallel \texttt{min} operation between the environment distance field and those of the humans at their predicted positions. An example composition is shown in Fig.~\ref{fig:composite_sdf}. 

In the case of only considering a single distance field of the environment for each observation update, i.e. no prediction, our method will not be of benefit since only one distance field computation is required. The benefit of our composition approach is apparent when multiple \textit{subsequent distance fields} are required for each environment update loop, such as our motion planning approach as detailed in Sec.~\ref{sec:rhapgpmp2} which considers multiple predicted distance fields of the environment for each update loop. As such, we perform benchmarking with respect to the calculation time for \textit{subsequent distance fields} against PBA. Hardware specifications used were: NVIDIA RTX 2060 GPU, 8-core Intel Core i7-9700 CPU @ \SI{4.50}{\giga\hertz} and \SI{2133}{\mega\hertz} DDR4~RAM.

Benchmarking results are shown in Fig.~\ref{fig:composite_sdf_benchmarking} where we provide comparisons both including and excluding the time to transfer distance fields from the GPU to the host device. Excluding the transfer times from GPU to host, Fig.~\ref{fig:composite_times_exc_transfer} shows our composite method to reduce computation time by \SIrange{89}{93}{\%}. Unfortunately, at the short timescales that we achieve, the transfer time becomes a dominant factor, accounting for over \SI{90}{\%} of the overall update time for the composite distance field. However, our composite method still provides a significant performance boost, even after accounting for the transfer time,  when compared to a full PBA calculation, cutting the computation time for the resultant distance field by \SIrange{40}{53}{\%}. 

\section{Human Trajectory Prediction}\label{sec:human_trajectory_prediction}
In indoor environments, a person typically moves towards an intended goal, such as a door to exit through or towards an object to pick up, rather than in a random manner. Therefore, the first component of our trajectory prediction module is \textit{intention recognition} in which we try to determine a person's intended goal.
\subsection{Intention Recognition}
\label{sec:int_rec}
Suppose there exists a set, $\mathcal{G}$, of $K$ possible goals for a person in the environment, $\bm{g}_k \in \mathcal{G}$, where a goal is represented by a 2D position vector $\bm{g}_k = [x_{g_k}, y_{g_k}]^T$. The purpose of the intention recognition module is first to recognise the possible goal locations in the environment that might be of interest to the person, i.e. $\mathcal{G}$, and secondly to identify which of these goals is a person's current intended goal, $\bm{g}$.

In practice, we believe that $\mathcal{G}$ can be learnt over time as objects and areas of interest are observed and identified. Similarly, we believe that we can infer possible goal locations by observing human motion data over time. To explore this idea, we consider a simple \textit{occupancy analysis} method. Given the recorded positions of a person over time, we discretise the positions across a 2D grid of arbitrary size and monitor the number of visitations for each cell, $n_{i}$, where the corresponding velocity is less than some threshold,~$v_{\text{thres}}$.
In the rest of this paper, we use $v_{\text{thres}} = \SI{0.3}{\metre\second^{-1}}$;
this value is much less than the average human walking velocity meaning that grid states which are frequently passed through will not be mislabeled as possible goals.
Using a threshold value closer to zero would lead to poor identification of goals where a person is not completely static but has slowed down, e.g. doors. Using the aforementioned frequency grid, where the most visited cell has $N_\text{max}$ visitations, we identify possible goal locations as those cells with $n_i > \frac{N_\text{max}}{2}$.
If there are multiple adjacent states identified as goals, we take the mean position of those states as a single goal location. By implicitly learning social context cues, our method can also learn additional goals that cannot easily be identified from semantics, e.g. particular gathering points without identifiable objects at those locations.
Note that the described goal estimation method can be supplemented with semantic information from the perception pipeline to use identified objects such as desks, sofas, and doors as possible goals, even if they were not visited during the observation time.

\begin{figure*}[tb]
          \subfloat[\label{fig:goals_a}]{
          \includegraphics[height=2.0in]{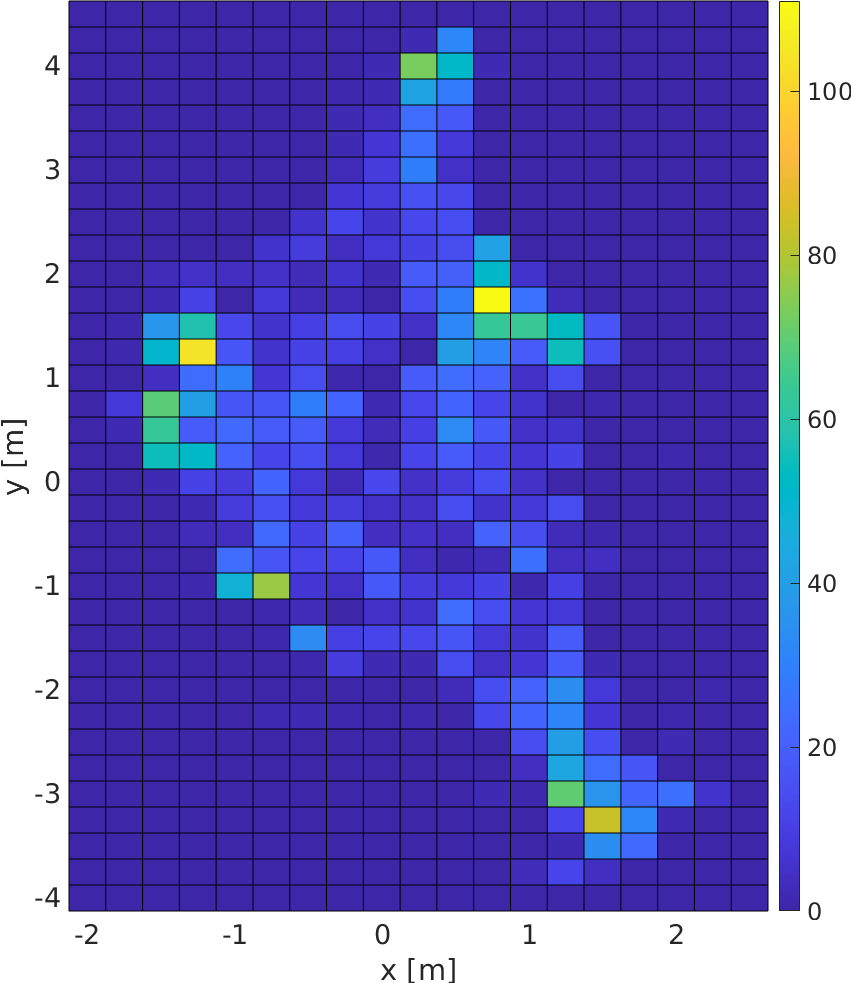}}   
          \hfill
          \subfloat[\label{fig:goals_b}]{
          \includegraphics[height=2.0in]{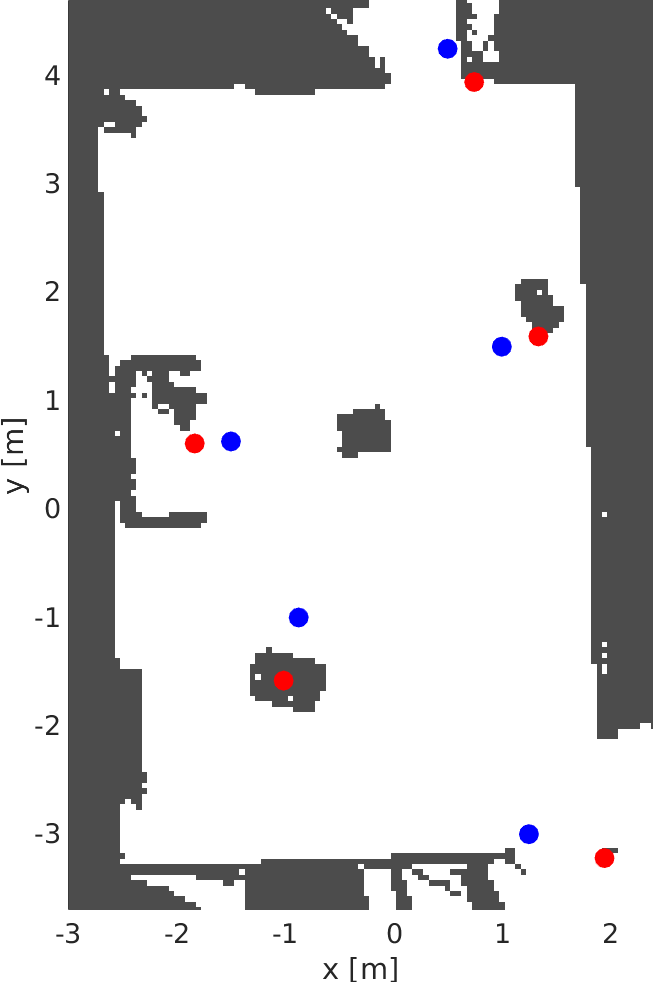}}   
          \hfill
          \subfloat[\label{fig:goals_c}]{
          \includegraphics[height=2.0in]{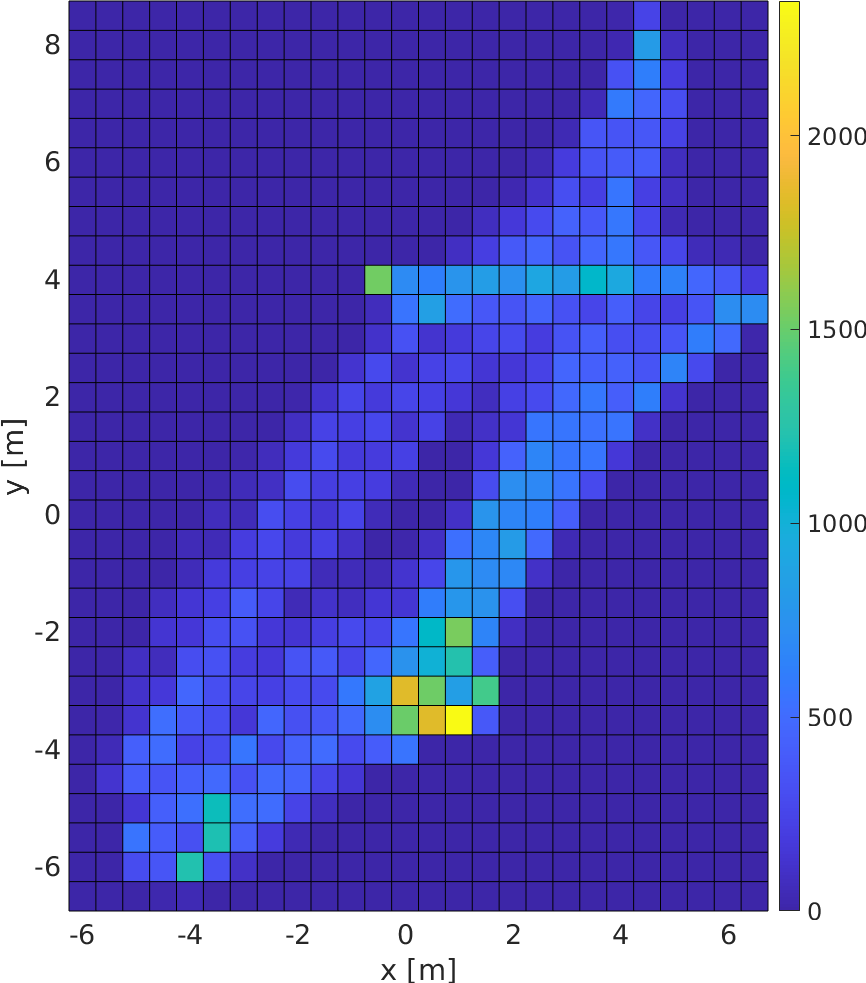}}   
          \hfill
          \subfloat[\label{fig:goals_d}]{
          \includegraphics[height=2.0in]{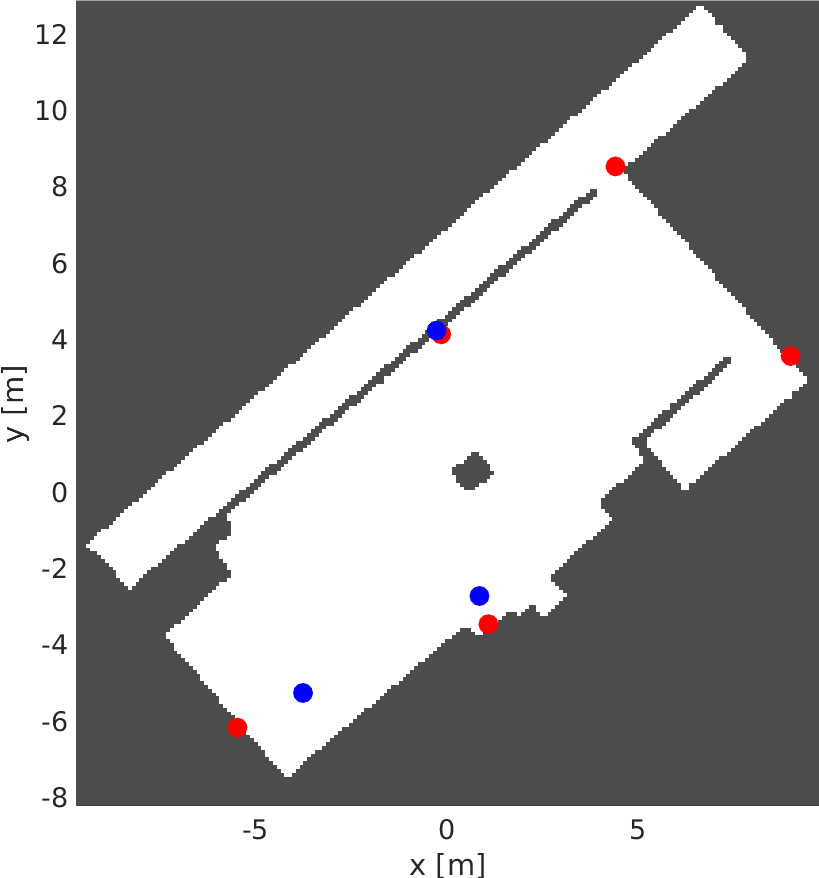}}   
          \hfill
\caption{By performing our occupancy analysis method on recorded motion data, we can estimate a person's possible goals. We demonstrate this technique using the Oxford-IHM (Figs. \ref{fig:goals_a} and \ref{fig:goals_b}) and TH{\"O}R datasets (Figs. \ref{fig:goals_c} and \ref{fig:goals_d}). We monitor the frequency of visitations to each grid state after applying a velocity threshold, resulting in the heatmaps shown. The most occupied states provide a reliable estimate of possible goal locations. Figures~\ref{fig:goals_b} and \ref{fig:goals_d} show maps of example environments from each dataset with the actual (blue) and estimated (red) goal locations.}
\label{fig:goals_occupancy}
\end{figure*}

We demonstrate our occupancy analysis method on both the Oxford-IHM and TH\"OR datasets; the results are shown in Fig.~\ref{fig:goals_occupancy} and indicate that the most commonly occupied grid states provide accurate estimates of the ground truth goal locations in each dataset.
For both datasets, we analysed segments $\sim 5$ minutes long. For the TH\"OR dataset, we tracked all the subjects marked as visitors~\citep{thorDataset2019}.
On the Oxford-IHM dataset, all ground truth goals were identified, although with an offset due to the ground truth goals being objects that humans maintain a distance from, e.g. a person sits in front of a desk rather than on it.
On the TH\"OR dataset, we identified three out of the five labelled goals; our method did not identify two of the goals for several reasons. Firstly, these goals were in areas with frequent motion capture track drops, a problem that does not occur with live robot sensor data. Secondly, these goals represented exit and entrance locations but without doors; as a result, people passed through without slowing down. One could argue that in this instance, these do not represent accurate goal locations since the people will continue to walk to their true goal locations.

For a given determination of $\mathcal{G}$, we want to determine the probability of each goal, $\bm{g}_k$, being the human's intended goal, $\bm{g}$, given that we have observed a history of the person's trajectory, $\bm{X}_h(t)$, where $t$ is the current time. We use $\bm{X}_h(t)$ to denote the matrix formed by the past $N$ vector measurements of the human's position, $\bm{x}_h(t_{i}) = [x_h(t_{i}), y_h(t_{i})]$ for times $t_i < t$.

By framing the human intention recognition problem in this probabilistic manner, we can use Bayes' rule to derive a posterior distribution for goal locations as
\begin{equation}
  p\big(\bm{G} | \bm{X}_h\big) \propto p(\bm{G}) p\big(\bm{X}_h | \bm{G}\big),
  \label{eq:bayes_intent}
\end{equation}
where $p(\bm{G})$ is a distribution that encodes prior knowledge of goal probabilities and $p\big(\bm{X}_h  | \bm{G}\big)$ is a conditional distribution that represents the \textit{likelihood} of the recorded past human trajectory for a set of a given goal.

If there is no prior knowledge about the probability distribution for goals, i.e. no goal visitation history, the prior $p(\bm{G})$ is set as the uniform distribution. On the other hand, if we have an observed (or pre-recorded) history of motion data and perform the occupancy grid analysis described previously, $p(\bm{G})$ can be set as a categorical distribution where the prior probability for each identified goal is proportional to its number of visitations, $N_k$:
\begin{equation}
  p(\bm{G} = \bm{g_k}) = \frac{N_k}{\sum_{l = 1}^K N_l}.
\end{equation}
Note that in practical applications, the prior goal distribution $p(\bm{G})$ can be initialised as a uniform distribution when a robot first begins to operate in an environment. As information about the environment and human movement is collected during operation, the prior distribution can be altered online after performing the grid occupancy analysis.

Lastly, we calculate the likelihood, $p\big(\bm{X}_h (t) | \bm{G}\big)$, i.e. given the human's recent history, what is the likelihood of each possible goal being the intended one?
Intuitively, a person is likely to look at the object they want to reach or move towards in the near future, i.e. the intended goal. As shown by \cite{kratzer2020prediction}, for a set of objects that represents possible goal locations in the environment, a person's gaze is a great predictor of intention. While we could build on this idea directly and use the difference in angle between a person's gaze and each object to determine the probability of each goal, determining a person's gaze in practice is challenging. Wearable gaze tracking equipment is shown to work very well~\citep{kratzer2020prediction}; however, it is impractical to assume that this is available in everyday applications such as in a household environment. On the other hand, alternative methods that try to estimate the human gaze from images \citep{kellnhofer2019gaze360, park2018deep} require significant computational resources to work in real-time and have degraded performance when a person is turned away from the robot.

For the aforementioned reasons, we use the human's estimated orientation, obtained from the history of positions, as a predictor of \textit{intent}. While the use of estimated orientation as a motion cue may not be as effective as using a person's gaze, it does not require observability of the human's face and can suffice when other motion cues are unavailable due to hardware constraints.
We estimate the human body orientation $\hat \theta_h(t_i)$ from the difference between two subsequent positions
\begin{equation}
  \hat \theta_h(t_i) = \arctan\bigg(\frac{y_h(t_i) - y_h(t_{i-1})}{x_h(t_i) - x_h(t_{i-1})}\bigg),
\end{equation}
where we use the `$h$' subscript to denote human positions. The relative orientation between a particular goal location and a person is then given by
\begin{equation}
  \delta\theta_{g_k}(t_i) = \arctan\bigg(\frac{y_{g_k} - y_h(t_i)}{x_{g_k} - x_h(t_i)}\bigg) - \hat \theta_h(t_i).
\end{equation}
In practice, there is likely to be noise in individual estimates of the orientation either due to sensor measurements or because a person may briefly look away from their goal without changing their intent. As a result, the relative orientation $\delta\theta_{g_k}(t_i)$ can quickly vary between subsequent timesteps without an actual change in the person's body motion.
Therefore we calculate the average over the past $N$ relative orientations
\begin{equation}
  \overline{\delta\theta_{g_k}}(t_{i}) = \frac{1}{N} \sum_{j=0}^{N-1} \delta\theta_{g_k}(t_{i-j}),
\end{equation}
We achieve more stable estimates for the relative orientation by aggregating the recent trajectory history. When the average relative orientation, $\overline{\delta\theta_{g_k}}(t_{i})$, is zero, it implies that a person is moving in the direction of the goal $\bm{g_k}$.
Conversely, when $\overline{\delta\theta_{g_k}}(t_{i})$ is equal to $\pi$, it implies that a person is moving directly away from the goal $\bm{g_k}$.
We thus formulate the likelihood $p\big(\bm{X}_h  | \bm{G} = \bm{g_k}\big)$ by calculating the softmax function of the average of past $N$ relative orientations~$\overline{\delta\theta_{g_k}}(t_{i})$
\begin{equation}
  p\big(\bm{X}_h  | \bm{G} = \bm{g_k}\big) = \frac{e^{\lambda\overline{\delta\theta_{g_k}}(t_{i})}}{ \sum_{l=1}^K e^{\lambda\overline{\delta\theta_{g_l}}(t_{i})}},
\end{equation}
where $\lambda$ is a constant that determines the sensitivity of the exponentiated cost. We use $\lambda = 1$ throughout the rest of this work.
Following Eq.~\ref{eq:bayes_intent}, the intended goal position is simply extracted by calculating the maximum a posteriori probability (MAP) estimate
\begin{equation}
  \hat{\bm{g}}_{\text{MAP}} = \underset{\bm{G} \in \mathcal{G}}{\text{arg\,max }}  p(\bm{G}) p\big(\bm{X}_h  | \bm{G}\big).
\end{equation}
The estimated goal, $\hat{\bm{g}}_{\text{MAP}}$, has the corresponding probability, $p\big(\bm{G}=\hat{\bm{g}}_{\text{MAP}} | \bm{X}_h \big)$, that represents how sure we are that the estimated goal is the intended one.
\subsection{Trajectory Optimisation}\label{sec:humantrajoptimisation}

Once we have determined a person's intended goal, we want to anticipate their motion towards it in order to safely steer the robot away from a person and avoid potential collisions.
We make several assumptions about human behaviour in our trajectory prediction method. 

Our first assumption is that a person, unobstructed by other factors, will move according to a constant-velocity kinematic model; a person walking directly towards a goal will tend to maintain the same velocity. 
While we could employ a higher-order kinematic model, such as constant-acceleration, it is unlikely that a person will quickly change their movement speed under normal circumstances, and so a constant-velocity motion model is sufficient for modelling human motion in open spaces~\citep{scholler2020constant}.
The second assumption that we make is that a person, unlike a moving inanimate object, is generally aware of obstacles in the environment and will try to avoid colliding with them. As such, a robot can use information that it has accumulated about the map of the environment as an environmental prior when predicting a person's trajectory.
Our third assumption is that a person is aware of robots operating in the environment and will tend to avoid the space that it occupies. 

Using these assumptions, we formulate the human trajectory prediction problem as non-linear trajectory optimisation; although primarily used for robot motion planning, in our implementation, we use GPMP2~\citep{gpmp2} as a state-of-the-art trajectory optimisation method.
Since human trajectory prediction and robot motion planning share similarities, we believe that GPMP2, with minor adaptations, is suitable for our prediction problem.

In GPMP2, the motion planning problem is framed as a probabilistic inference problem whereby the aim is to formulate the posterior density of a trajectory and solve for the maximum a posteriori (MAP) estimator, just as we did in the previous section. Using Bayes' rule, the posterior distribution of a trajectory, $\bm{x}$, given the likelihood on a collection of events, $\bm{e}$, is given by
\begin{align}\label{eq:probinf}
p(\bm{x} | \bm{e}) \propto p(\bm{x}) p(\bm{e}|\bm{x}),
\end{align}
where $p(\bm{x})$ represents the prior that encourages trajectory smoothness, while $p(\bm{e}| \bm{x})$ represents the probability of the events $\bm{e}$ occurring given $\bm{x}$. In the case of motion planning, $\bm{e}$ corresponds to binary events that a trajectory $\bm{x}$ is collision-free at a particular state.

In GPMP2, robot trajectories are represented as samples from a continuous-time Gaussian Process (GP), $\bm{x}(t) \sim \mathcal{GP}(\bm{\mu}(t), \bm{\mathcal{K}}(t, t^{\prime}))$, where $\bm{\mu}(t)$ is the vector-valued mean trajectory and $\bm{\mathcal{K}}(t, t^{\prime})$ is the matrix-valued covariance. By carefully choosing a structured kernel, one can show that the resultant precision matrix is exactly-sparse \citep{gpmp2}. Consequently, \cite{gpmp2} show that the probabilistic inference problem in Eq.~\ref{eq:probinf} can be efficiently solved on a factor graph. 

We adopt a similar factor graph formulation and adapt it for human trajectory prediction. Using a structured kernel as in \cite{gpmp2}, the prior and the likelihood functions can be written as a product of functions
\begin{equation}
\begin{split}
  p(\bm{x}_h) p(\bm{e} | \bm{x}_h) & \propto f^{p r i o r}\left(\bm{{X}}_{h}\right) f^{l i k e}\left(\bm{{X}}_{h}\right) \\
  & = \prod_{i} f_{i}\left(\bm{{X}}_{h, i}\right) \\
\end{split}
\label{eq:factorization}
\end{equation}
 where $\bm{X}_h = \{ \bm{x}_{h,0}, \dots, \bm{x}_{h,N} \}$ represents the set of future human positions along the predicted trajectory.
 The \textit{factors} $\mathcal{F} = \{ f_0, \dots, f_M \}$ are functions that act on variable subsets of the trajectory. As shown by \cite{kschischang2001factor}, the posterior distribution can be represented by a bipartite factor graph $G = \{\bm{X}, \mathcal{F}, \mathcal{E}\}$, where $\mathcal{E}$ is the set of \textit{edges} that connect \textit{variable} and \textit{factor} nodes.

To encourage \textit{smoothness} in our predicted human trajectories and account for the tendency to move according to a constant-velocity motion model, we adopt the GP prior proposed in \cite{gpmp2},
\begin{equation}
  p(\bm{x}_h) \propto \text{exp}\{ - \frac{1}{2} \norm{\bm{x}_h - \bm{\mu}_h }_{\bm{\mathcal{K}}_h}^2 \},
  \label{eq:gpprior}
\end{equation}
given in terms of the mean trajectory ${\bm{\mu}}_h$ and covariance ${\bm{\mathcal{K}}}_h$.
We initialise the mean as a constant-velocity straight line, while the covariance is obtained by solving the Linear Time-Varying Stochastic Differential Equation (LTV-SDE) with constant-velocity model system matrices, as in~\cite{gpmp2}. Due to the structured kernel choice, this GP prior has a Markovian structure; as such, it can be written as a product of GP prior factors that depend only on two neighbouring states, $f^{gp}(\bm{x}_{h,i}, \bm{x}_{h,i+1})$.

In addition to GP priors that describe how our trajectory behaves, we want to impose knowledge of a person's start and intended goal states. In the context of the human trajectory prediction, the start state is the current position of a person in the environment, while the intended goal state is obtained by our intention recognition method described in Sec.~\ref{sec:int_rec}.
We encode start and goal states by using the following factors:
\begin{equation}
  f^{s t a r t}{\left(\bm{x}_{h, 0}\right)} = \text{exp}\{ - \frac{1}{2} \norm{\bm{x}_{h, 0} - \bm{x}_{current} }_{\bm{\Sigma}_{h, 0}}^2 \},
\end{equation}
\begin{equation}
  f^{g o a l}{\left(\bm{x}_{h, N}\right)} = \text{exp}\{ - \frac{1}{2} \norm{\bm{x}_{h, N} - \bm{x}_{goal} }_{\bm{\Sigma}_{h, N}}^2 \},
\end{equation}
where $N$ represents the final support state of the trajectory. $\bm{\Sigma}_{h, 0}$ and $\bm{\Sigma}_{h, N}$ are the isotropic covariance matrices for the start and goal states. Smaller values along the diagonals of these matrices result in higher costs for deviating from the specified start and goal states, encouraging the optimised human trajectory prediction to adhere to the start and goal state constraints.

The motion prior part of the factorisation in Eq.~\ref{eq:factorization} thus becomes a product of factors that takes into account the current position of a person, their intended goal and the constant-velocity movement assumption
\begin{multline}
 f^{p r i o r}(\bm{X}_h) = \\ f^{s t a r t}{\left(\bm{x}_{h, 0}\right)} f^{g o a l}{\left(\bm{x}_{h, N}\right)} \prod_{i=0}^{N-1} f^{gp}{\left(\bm{x}_{h,i},\bm{x}_{h,i+1}\right)}.
\end{multline}

The remaining product of factors represents the likelihood $f^{l i k e}(\bm{X}_h)$ and encodes all other state-dependent costs and constraints. In the case of human trajectory prediction, we partition it into separate factors that encode collision avoidance with respect to the environment, $f^{o b s}$, and collision avoidance with respect to the moving robot, $f^{r o b o t}$. The likelihood thus becomes
\begin{equation}
 f^{l i k e}(\bm{X}_h) = \prod_{i=1}^{N-1} f_{i}^{o b s}\left(\bm{x}_{h,i}\right) f_{i}^{r o b o t}\left(\bm{x}_{h,i}\right).
\end{equation}

For environment collision avoidance factors, we adopt the formulation from GPMP2~\citep{gpmp2} which uses a hinge loss function on the Euclidean distance field of the environment to penalise states that are close to obstacles. As described in Sec.~\ref{sec:rhapgpmp2}, in practice, this distance field is provided by the perception part of our pipeline and updated online; this is in contrast to previous works which pre-compute it \cite{gpmp2, Park2012}.

For the robot avoidance factor, $f^{r o b o t}$, we propose
\begin{equation}
f_{i}^{r o b o t}\left(\bm{x}_{h,i}\right) = \text{exp}\{ - \frac{1}{2} \norm{\bm{h}({\bm{x}_{h,i})}}_{\bm{\Sigma}_r}^2 \},
\label{eq:robot_avoidance}
\end{equation}
where $\bm{h}(\bm{x}_{h,i})$ is the hinge loss function of the distance between a person and the robot at the current position, $\bm{x}_r$. The hinge loss function is defined as
\begin{equation}
\bm{h}(\bm{x}_{h,i}) = 
\begin{cases}
\centering
\mathcal{\varepsilon}_{r} - \norm{\bm{x}_{h,i} - \bm{x}_r}_2 &\mbox{ if } \norm{\bm{x}_{h,i} - \bm{x}_r}_2 \leq \varepsilon_{r} \\
\hfil  0  &\mbox{ if } \norm{\bm{x}_{h,i} - \bm{x}_r}_2 > \varepsilon_{r}
\end{cases}.
\end{equation}
$\varepsilon_{r}$ is a tolerance parameter of our formulation which indicates how close a person is likely to get to a robot before altering their trajectory to avoid collision.
If a person is sufficiently far away from the robot, we assume that they will not change their behaviour.
However, if the robot comes within the \textit{safety distance}, $\varepsilon_{r}$, our assumption is that a person will change their behaviour to move away from the robot and avoid collision.

The complete factor graph that we propose for human trajectory prediction can be written as
\begin{equation}
  p(\bm{x_h} | \bm{e}) \propto f^{s t a r t} f^{g o a l} \prod_{i=0}^{N-1} f^{gp}_{i, i+1} \prod_{i=1}^{N-1} f^{obs}_i f^{robot}_i,
\end{equation}
If we cannot determine a person's intended goal, for instance, if we do not have a set of possible goals or the estimated goal's probability is low, we can omit the goal prior factor. Our proposed prediction method will then work as a constant-velocity model that considers collisions via the robot and obstacle factors.
We perform inference on the factor graph using the Levenberg-Marquardt optimisation method implemented in GTSAM~\cite{dellaert2012factor} with an initial damping parameter of $0.01$. 
\subsection{Evaluation}\label{sec:prediction_eval}

We evaluate the proposed trajectory prediction method on our human motion dataset described in Sec.~\ref{sec:dataset} and on the TH{\"O}R public dataset of human motion trajectories \citep{thorDataset2019}.
On both datasets, we predict over four different prediction horizons: \SI{1.6}{\second}, \SI{3.2}{\second}, \SI{4.8}{\second}, and \SI{8.0}{\second}.
These prediction horizons cover both short-term and long-term human motion prediction and have previously been used in multiple evaluation pipelines, including the ATLAS benchmark~\citep{Rudenko2021}.
We thus use them for retaining consistency with the existing body of work in human motion prediction evaluation.
\subsubsection{Oxford-IHM Dataset}\label{sec:oxf-evaluation}
Our dataset comprises three different sensor measurements on which we evaluate trajectory prediction performance: the Vicon motion capture data, a static RGB-D camera and an HSR's head-mounted RGB-D camera.
The Vicon motion capture data serves as the ground truth against which we compare our predicted trajectories for each type of sensor data.
Evaluation of prediction methodologies on the motion capture data gives an indication of their potential performance when given accurate, high-frequency streams of data for the robot's pose, person's pose and goal locations.
On the other hand, evaluation of each method on the data obtained using static and robot-mounted RGB-D cameras indicates their respective prediction performance when operating in real-world environments. Predictions on these data sources account for errors in human position estimation arising from factors such as measurement noise, misdetections, and occlusions.

Further, we compare the performance of our proposed method on the motion capture data with two ablations: (1) without the intention recognition proposed in Sec.~\ref{sec:int_rec},  (2) without the robot avoidance factor proposed in Eq.~\eqref{eq:robot_avoidance}.
These two ablation studies enable us to assess the respective impact of the two features on human motion prediction performance.
For each data source, we compare the performance of our proposed method against two baselines: (1)~Constant Velocity Model (CVM), (2) Linear Velocity Model (LVM), similar to the ATLAS benchmark~\citep{Rudenko2021}.
The \textit{CVM} generates predictions by forward propagating the velocity of the person's last observed state, while the \textit{LVM} model generates predictions by forward propagating the observed average velocity of the person. 
We evaluate trajectory prediction performance using commonly used geometric metrics: Average Displacement Error (ADE) and Final Displacement Error (FDE)~\citep{rudenko2020human}. ADE measures the average error across predicted trajectories and the ground truth trajectory, while FDE measures the error of the final predicted point.
Since motion capture and RGB~-D measurements are inherently asynchronous, for evaluation on RGB-D data sources, we use GP interpolation~\citep{gpmp2} between states that are temporally closest to the ground truth measurements. For the two baselines methods, we use linear interpolation.

\begin{figure}[t]
\centering
\includegraphics[width=\columnwidth]{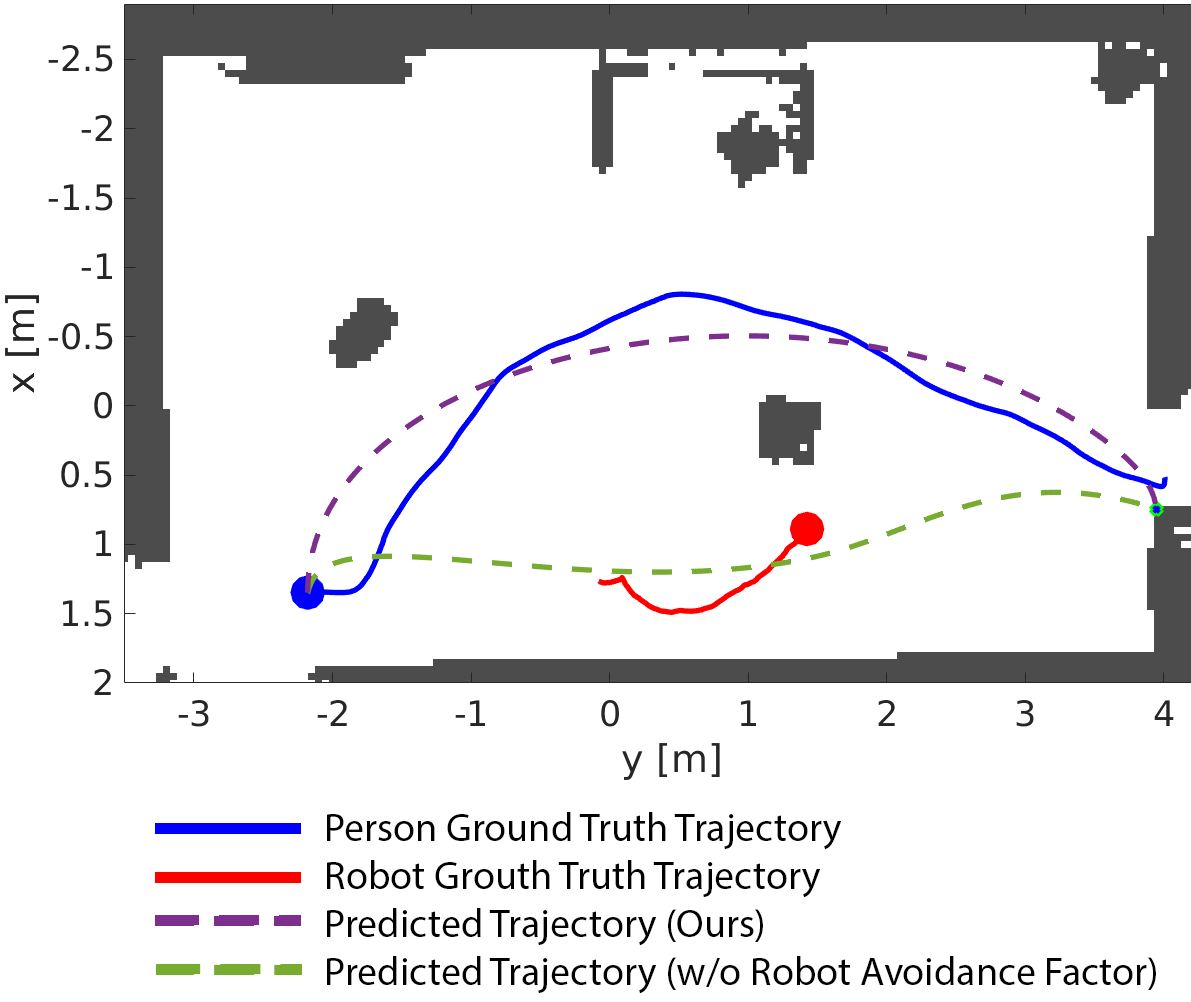}
\caption{An instance in the Oxford-IHM dataset that highlights the impact of using a robot avoidance factor. Without the proposed robot avoidance factor, the predicted human trajectory significantly deviates from the ground truth and collides with the robot.}
\label{fig:robot_factor}
\end{figure}

The results of evaluating across all 12 runs in the Oxford-IHM dataset, are shown in Tables \ref{table:custom_benchmark_ade} and \ref{table:custom_benchmark_fde}.
From the results, our proposed method outperforms the baseline methods for each type of sensor data and every prediction horizon.
For the shortest prediction horizon (\SI{1.6}{\second}), our proposed method performs similarly to the \textit{CVM} baseline; this is expected since our proposed method has a smoothness factor that is initialised with a constant-velocity motion model.
For short prediction horizons, goal and environmental factors have a marginal impact on human motion in absolute terms, meaning that simple kinematic models can often suffice.
However, as we predict over longer horizons, our proposed method significantly outperforms the baselines, in line with our expectations, since the \textit{CVM} and \textit{LVM} models do not predict goal-oriented behaviour and disregard environmental cues.
Without intent recognition, our method achieves similar performance to the \textit{CVM} and performs significantly worse than our complete proposed method, demonstrating the importance of intention recognition for human motion prediction in indoor environments.
Our proposed method performs similarly with and without using the robot avoidance factor, with only minor improvements being achieved with the robot avoidance factor.
However, we believe that this factor may become more significant when operating in cluttered environments since it achieves better prediction in specific cases, for example, when the robot blocks a direct path to the intended goal.
Figure~\ref{fig:robot_factor} shows an instance of such a situation in the Oxford-IHM dataset.

Using data from the static and robot-mounted RGB~-D cameras, we see similar performance trends to those achieved using motion capture measurements, albeit with a greater error. However, the dominant source of this error arises from our position estimation approach applied to the RGB-D images. By comparing the position estimates obtained using our image processing with ground truth measurements, we observe average position estimation errors of \SI{15.9}{\centi\metre} and \SI{30.4}{\centi\metre} respectively for the HSR RGB-D and Static RGB-D data sources. These results suggest that by further improving the position estimation method, similar trajectory prediction performance can be achieved on RGB-D sources to that achieved using high-frequency motion capture.
While the static sensor had fewer measurement drops due to having the whole environment in its field of view, the robot-mounted camera achieved better performance since the robot was usually closer to the person and thus more in accordance with the camera's recommended `distance of use' for the camera.
The results indicate that our proposed human trajectory prediction method works effectively with live sensor data and can be integrated within our proposed perception and motion planning framework as described in previous sections.

\begin{table*}[t]
% \small
\begin{tabularx}{\linewidth}{| c | C | C | C | C | C |}
\multicolumn{2}{ c }{}&\multicolumn{4}{ c }{} \\ \cline{3-6}
\multicolumn{2}{ c |}{}&\multicolumn{4}{ c |}{Prediction horizon (\si{\second})}\\
\hline
 & Method & 1.6 & 3.2 & 4.8 & 8.0 \\
\hline
\multirow{5}{*}{Vicon} 
& LVM & $0.5 \pm 0.02$ & $1.08 \pm 0.04$ & $1.7 \pm 0.05$ & $2.98 \pm 0.07$\\ 
& CVM & $\bm{0.28 \pm 0.03}$ & $0.75 \pm 0.04$ & $1.35 \pm 0.07$ & $2.75 \pm 0.11$\\ 
& Ours & $\bm{0.28 \pm 0.02}$ & $\bm{0.66 \pm 0.06}$ & $\bm{0.99 \pm 0.09}$ & $\bm{1.54 \pm 0.11}$\\ 
& Ours w/o Factor & $\bm{0.28 \pm 0.03}$ & $0.68 \pm 0.07$ & $1.04 \pm 0.09$ & $1.59 \pm 0.12$\\
& Ours w/o Intent & $\bm{0.28 \pm 0.03}$ & $0.74 \pm 0.05$ & $1.25 \pm 0.08$ & $2.54 \pm 0.12$\\ 
\hline
\multirow{3}{*}{Static RGB-D} 
& LVM & $0.78 \pm 0.03$ & $1.38 \pm 0.06$ & $2.02 \pm 0.09$ & $3.28 \pm 0.1$\\ 
& CVM & $\bm{0.59 \pm 0.03}$ & $1.03 \pm 0.06$ & $1.65 \pm 0.1$ & $3.03 \pm 0.13$\\ 
& Ours & $\bm{0.59 \pm 0.04}$ & $\bm{0.95 \pm 0.08}$ & $\bm{1.13 \pm 0.1}$ & $\bm{1.86 \pm 0.14}$\\ 
\hline
\multirow{3}{*}{HSR RGB-D} 
& LVM & $0.66 \pm 0.03$ & $1.26 \pm 0.06$ & $1.89 \pm 0.09$ & $3.12 \pm 0.09$\\ 
& CVM & $0.44 \pm 0.03$ & $0.91 \pm 0.05$ & $1.54 \pm 0.08$ & $2.99 \pm 0.10$\\ 
& Ours & $\bm{0.43 \pm 0.03}$ & $\bm{0.81 \pm 0.07}$ & $\bm{1.13 \pm 0.1}$ & $\bm{1.71 \pm 0.13}$\\ 
\hline
\end{tabularx}
\caption{Average Displacement Error (ADE) on the Oxford-IHM dataset}
\label{table:custom_benchmark_ade}
\end{table*}

\begin{table*}[t]
% \small
\begin{tabularx}{\linewidth}{| c | C | C | C | C | C |}
\multicolumn{2}{ c }{}&\multicolumn{4}{ c }{} \\ \cline{3-6}
\multicolumn{2}{ c |}{}&\multicolumn{4}{ c |}{Prediction horizon (\si{\second})}\\
\hline
 & Method & 1.6 & 3.2 & 4.8 & 8.0 \\
\hline
\multirow{5}{*}{Vicon} 
& LVM & $1.03 \pm 0.03$ & $2.29 \pm 0.07$ & $3.6 \pm 0.09$ & $6.09 \pm 0.18$\\ 
& CVM & $\bm{0.64 \pm 0.04}$ & $1.82 \pm 0.08$ & $3.26 \pm 0.14$ & $6.41 \pm 0.2$\\ 
& Ours & $0.65 \pm 0.04$ & $\bm{1.36 \pm 0.18}$ & $\bm{1.98 \pm 0.14}$ & $\bm{2.88 \pm 0.15}$\\ 
& Ours w/o Factor & $0.65 \pm 0.04$ & $1.38 \pm 0.18$ & $2.02 \pm 0.14$ & $2.94 \pm 0.16$\\
& Ours w/o Intent & $0.65 \pm 0.04$ & $1.82 \pm 0.08$ & $3.2 \pm 0.14$ & $6.11 \pm 0.21$\\
\hline
\multirow{3}{*}{Static RGB-D} 
& LVM & $1.34 \pm 0.04$ & $2.6 \pm 0.08$ & $3.88 \pm 0.11$ & $6.38 \pm 0.2$\\ 
& CVM & $\bm{0.92 \pm 0.05}$ & $2.1 \pm 0.07$ & $3.54 \pm 0.13$ & $6.59 \pm 0.22$\\ 
& Ours & $\bm{0.92 \pm 0.05}$ & $\bm{1.63 \pm 0.14}$ & $\bm{2.24 \pm 0.16}$ & $\bm{3.13 \pm 0.18}$\\ 
\hline
\multirow{3}{*}{HSR RGB-D} 
& LVM & $1.18 \pm 0.04$ & $2.45 \pm 0.09$ & $3.75 \pm 0.09$ & $6.17 \pm 0.18$\\ 
& CVM & $\bm{0.80 \pm 0.04}$ & $1.98 \pm 0.1$ & $3.41 \pm 0.16$ & $6.51 \pm 0.16$\\ 
& Ours & $0.81 \pm 0.04$ & $\bm{1.42 \pm 0.13}$ & $\bm{2.12 \pm 0.18}$ & $\bm{3.02 \pm 0.18}$\\ 
\hline
\end{tabularx}
\caption{Final Displacement Error (FDE) on the Oxford-IHM dataset}
\label{table:custom_benchmark_fde}
\end{table*}

\subsubsection{TH{\"O}R dataset}
We further benchmark our proposed method trajectory prediction on the TH{\"O}R dataset~\citep{thorDataset2019} in which ten human subjects are tracked in an indoor environment with static obstacles and perform four different social roles that imitate typical activities found in populated spaces (e.g. offices). Enacting these roles results in various motion patterns, and nine out of the ten subjects exhibit goal-oriented behaviour. The dataset includes five labelled goals with known ground truth positions.
The motion capture data provides ground truth trajectories against which we compare our predictions.

We use the ATLAS benchmark~\citep{Rudenko2021} to compare the performance of our proposed method against five different methods, including two baselines (\textit{CVM} and \textit{LVM}). The other three methods are local interaction models, namely the Social force model (\textit{Sof}) \citep{helbing1995social} and its two predictive extensions \textit{Zan} \citep{zanlungo2011social} and \textit{Kara} \citep{karamouzas2009predictive}).
These models consider that multiple people are moving in the same environment and will anticipate and evade collisions with each other. 
As with the Oxford-IHM dataset, we evaluate trajectory prediction performance using the ADE and FDE.

The results of benchmarking for all subjects, across all four runs of the TH{\"O}R  \textit{One obstacle} experiment, are shown in Table \ref{table:atlas}.
In contrast to the Oxford-IHM dataset, the TH{\"O}R dataset features fewer obstacles and a larger environment, resulting in more straight-line trajectories with constant velocity. Consequently, we achieve better performance than on the Oxford-IHM dataset.
Our method is shown to outperform the baseline methods for all prediction horizons. While the \textit{CVM} baseline achieves a similar level of performance on the shortest prediction horizon, our proposed method significantly outperforms on longer horizons for the reasons explained in Sec.~\ref{sec:oxf-evaluation} and in line with our expectations.
Our method marginally outperformed the local interaction models (\textit{Sof}, \textit{Zan} and \textit{Kara}) on the ADE metric, but was marginally worse on the FDE metric.

The relatively high standard deviations of the proposed method can be explained by the different social roles assigned to subjects on the TH{\"O}R dataset. For the \textit{Lab Worker} and \textit{Utility Worker} roles, the proposed method achieves superior performance in both ADE and FDE because these roles operate in a very goal-oriented manner with no social interactions.
In contrast, subjects in the social role, \textit{Visitor}, exhibited behaviours not currently modelled by our method, such as slowing down to interact with other people. For these roles, our proposed method had performed worse than the local interaction models which consider the social component of human behaviour. By not accounting for people slowing down, our proposed method often predicted people travelling further than the ground truth, mainly affecting the FDE performance.

\begin{table*}[t]
\begin{tabularx}{\linewidth}{|c|c|C|C|C|C|}
\hline & \multicolumn{5}{|c|}{ Prediction horizon (\si{\second})} \\
\cline { 2 - 6 } & Method & $1.6 $ & $3.2 $ & $4.8 $ & $8.0$ \\
\hline \multirow{5}{*}{ADE} & CVM & $\bm{0.15 \pm 0.09}$ & $0.38 \pm 0.24$ & $0.71 \pm 0.45$ & $1.51 \pm 0.91$ \\
 & LIN & $0.29 \pm 0.18$ & $0.60 \pm 0.38$ & $0.99 \pm 0.63$ & $1.84 \pm 1.08$ \\
 & Sof & $0.18 \pm 0.10$ & $0.36 \pm 0.20$ & $0.60 \pm 0.35$ & $1.13 \pm 0.67$ \\
 & Zan & $\bm{0.15 \pm 0.09}$ & $0.34 \pm 0.20$ & $0.59 \pm 0.36$ & $1.16 \pm 0.70$ \\
 & Kara & $0.16 \pm 0.08$ & $0.35 \pm 0.19$ & $0.60 \pm 0.36$ & $1.16 \pm 0.69$ \\
 & Ours & $\bm{0.15 \pm 0.09}$ & $\bm{0.33 \pm 0.26}$ & $\bm{0.57 \pm 0.41}$ & $\bm{1.12 \pm 0.83}$ \\
 \hline
\hline
 \multirow{5}{*}{FDE} & CVM & $0.28 \pm 0.18$ & $0.86 \pm 0.54$ & $1.64 \pm 1.05$ & $3.54 \pm 2.11$ \\
 & LIN & $0.49 \pm 0.31$ & $1.20 \pm 0.75$ & $2.07 \pm 1.30$ & $3.97 \pm 2.27$ \\
 & Sof & $0.29 \pm 0.16$ & $\bm{0.72 \pm 0.42}$ & $\bm{1.27 \pm 0.79}$ & $\bm{2.48 \pm 1.54}$ \\
 & Zan & $\bm{0.26 \pm 0.16}$ & $\bm{0.72 \pm 0.43}$ & $1.31 \pm 0.82$ & $2.62 \pm 1.61$ \\
 & Kara & $0.28 \pm 0.15$ & $0.73 \pm 0.42$ & $1.31 \pm 0.82$ & $2.59 \pm 1.59$ \\
 & Ours & $0.28 \pm 0.17$ & $0.78 \pm 0.64$ & $1.41 \pm 0.99$ & $2.98 \pm 1.89$ \\
 \hline
\end{tabularx}
\caption{ADE and FDE on the TH\"OR dataset}
\label{table:atlas}
\end{table*}

\subsubsection{Parameters}

Since we use GPMP2 as the backbone for trajectory optimisation, our trajectory prediction method depends on a similar set of parameters.
Its parameter $Q_c$ specifies the uncertainty in the prior distribution and determines how heavily states are penalised for deviating away from the mean. 
$\Sigma_{obs}$ represents the obstacle cost weight with smaller values more strongly penalising collisions with obstacles. 
Since we introduce the robot avoidance factor in Eq.~\ref{eq:robot_avoidance}, we have the additional parameter, $\Sigma_{r}$, that we set to the same value as $\Sigma_{obs}$ throughout this paper, equally penalising collisions with the robot and the static environment.
The parameter $\varepsilon$ represents a \textit{safety distance} from static obstacles. For larger values of $\varepsilon$, optimised trajectories will deviate more from a straight line to maintain a larger distance from obstacles. 
The proposed robot avoidance factor has a similar parameter, $\varepsilon_{r}$, indicating a desired \textit{safety distance} from the robot.

For our evaluation, we performed a grid search over parameters $Q_c$ and $\Sigma_{obs}$ for the Oxford-IHM and TH\"OR datasets.
We found that good trajectory performance was achieved for parameters in the following ranges; $Q_c \in [0.01, 0.5]$ and $\Sigma_{obs} \in [0.02, 0.3]$.
We used $Q_c = 0.2$ on Oxford-IHM and $Q_c = 0.05$ on the TH\"OR dataset.
Since ground truth human trajectories on the Oxford-IHM dataset were less smooth than on TH\"OR due to its environment being smaller and more cluttered, a larger value of $Q_c$ was needed to better account for obstacle avoidance in tight spaces.
On both datasets we used $\Sigma_{obs} = 0.1$ and $\varepsilon = 0.4$, resulting in collision-free trajectories for most predictions. Note that $\varepsilon$ also considers the radius of a person since the obstacle cost looks at the distance between the \textit{centre} of a person and the nearest obstacle.
The \textit{safety distance} from the robot was set as $\varepsilon_{r} = 0.8$, double the distance from static obstacles because we account for the robot's radius and also that a person is likely to move farther from a moving robot than a static obstacle.

Due to the underlying continuous-time trajectory representation, we must define the total duration of a trajectory, corresponding to an estimation of the time required for a person to reach their intended goal.
We estimate this time by calculating the distance between the person's current position and their intended goal and dividing it by their current velocity.
For estimated times shorter than the prediction horizon used for evaluation, we use the goal state to predict a person's position for timestamps after the estimated trajectory time.

\section{Receding Horizon And Predictive Gaussian Process Motion Planner 2}\label{sec:rhapgpmp2}

\begin{figure*}[t]
\centering
\includegraphics[draft=false,width=2.0\columnwidth]{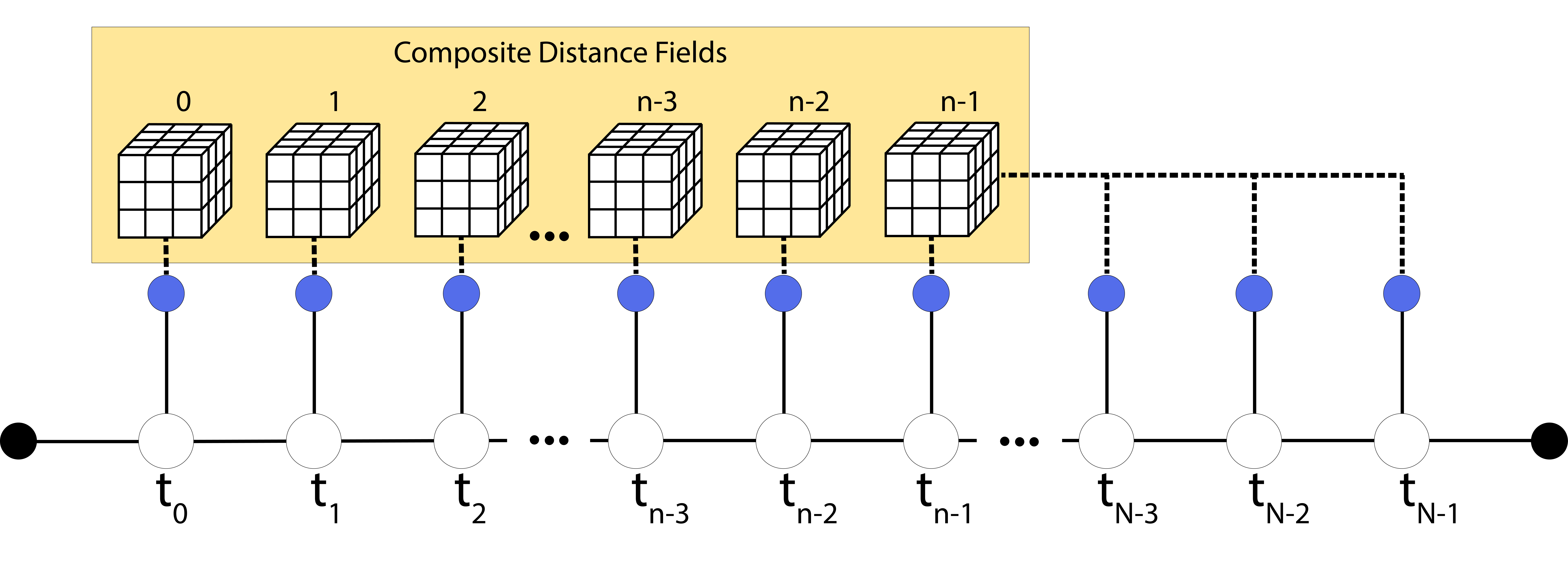}
\caption{The assignment of composite distance fields to the obstacle factors (blue) in RHAP-GPMP2. Given a long time horizon of $N$ timesteps, we assign independent distance fields to the first $n$ timesteps, where $n$ is our dynamic obstacle prediction horizon. For time-indexed obstacle factors greater than $n$, we assign the $n^{\text{th}}$ distance field.}
\label{fig:factor_graph}
\end{figure*}

This section describes how we integrate the methods and concepts discussed in previous sections within a single framework to be deployed on a physical robot. We build upon the integrated perception and motion planning framework described in \cite{FineanIros2021}; we use the GPU-Voxels framework to maintain a voxelmap of the scene and compute distance fields \citep{GPUVoxels, GPUVoxelsMobile}, while motion planning is performed using GPMP2 in a receding-horizon manner. However, we introduce several extensions.

Firstly, as discussed in Sec.~\ref{sec:frameworkoverview} and Sec.~\ref{sec:image_segmentation}, we introduce image segmentation to remove dynamic obstacles prior to generating pointclouds for integration into the maintained voxelmap of the static scene. At this point, the voxelmap can undergo further filtering if necessary. In the presence of dynamic obstacles, we found it beneficial to filter out voxels that have fewer than five connected voxels; this reduced the instances of spurious voxels being designated as occupied in the voxelmap of the static scene.

Secondly, we propose the Receding Horizon And Predictive Gaussian Process Motion Planner~2 (RHAP-GPMP2). As described in Sec.~\ref{sec:humantrajoptimisation}, GPMP2 formulates the motion planning problem as probabilistic inference on a factor graph. In RHAP-GPMP2, we continuously monitor the validity of the current trajectory, re-estimate the expected time-to-goal, and re-optimise trajectories to re-evaluate their cost as we observe the environment. If the current trajectory becomes invalid or a re-optimised trajectory significantly lowers the cost, we generate a new factor graph for trajectory optimisation. We use a straight-line trajectory initialisation for the first optimisation and in recovery behaviours; otherwise, we re-use and re-optimise the previously planned trajectory to maintain smoothly executed trajectories. In previous work, we used a singular voxelmap that is maintained, converted to a distance field, and sent to all obstacle factors within the factor graph used for motion planning. However, RHAP-GPMP2 builds upon the concepts presented in \cite{FineanComposite2020} and extends the motion planning work to time-configuration space planning. To achieve this, we maintain: 
\begin{enumerate}
    \item A static voxelmap of the scene
    \item A distance field (static or maintained) for each dynamic obstacle (discussed in Sec.~\ref{sec:sdfs})
    \item A sequence of $n$ composite distance fields.
\end{enumerate}
The variable $n$ is determined by how far into the future we wish to incorporate predicted positions for moving objects in the scene. In this work, we use a time-discretisation between factor graph support states of \SI{0.5}{\second} and so choose a value of $n = 20$, corresponding to a time horizon of \SI{10}{\second}. Each time-indexed obstacle factor in the factor graph is associated with a corresponding time-indexed composite distance field. For time-indices greater than $n$, we assign the composite distance field for time index $n$. Distance field assignment for RHAP-GPMP2 is illustrated in Fig.~\ref{fig:factor_graph}.

During each update loop, the static voxelmap is updated using the latest observations of the scene (with the dilated dynamic obstacle masks removed) and composite distance fields are generated using the latest trajectory predictions for dynamic obstacles in the scene, as predicted by our trajectory prediction module described in Sec.~\ref{sec:frameworkoverview}, Sec.~\ref{sec:int_rec}, and Sec.~\ref{sec:humantrajoptimisation}. To integrate the human prediction module, we additionally calculate a 2D EDT of the environment by collapsing the maintained 3D voxel grid to 2D and using PBA to calculate the corresponding EDT on the GPU. The EDT is then transferred to the CPU for use in the trajectory prediction module. 

\section{Live Hardware Experiments}
\begin{figure*}[tb]
        \centering
          \subfloat[Start\label{fig:change_places_t0}]{
          \includegraphics[width=0.66\columnwidth]{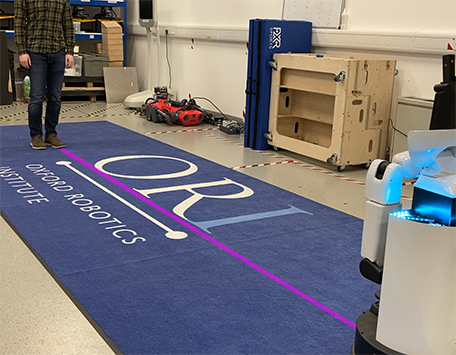}}   
          \hfill
          \subfloat[\textit{No Prediction} results in collision\label{fig:change_places_t55}]{
          \includegraphics[width=0.66\columnwidth]{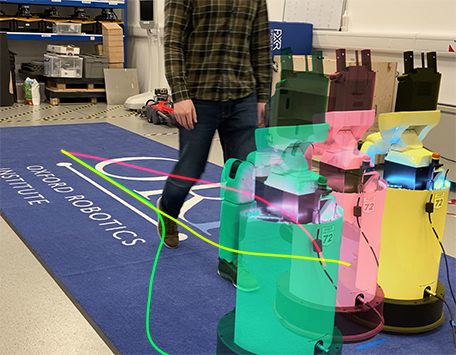}} 
          \hfill
          \subfloat[Both predictive methods succeed\label{fig:change_places_t62}]{
          \includegraphics[width=0.66\columnwidth]{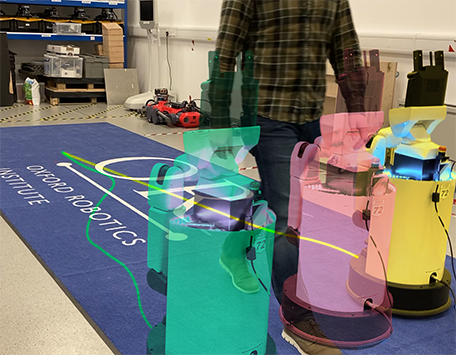}}  
\caption{\textit{Change of Places} -- the robot is tasked with a base-only goal in front of a person. During execution, the person walks towards a goal behind the robot, requiring the robot to react and move out of the way. All trajectories initially follow a straight line (purple), but as the person approaches, the replanned trajectories for each method diverge as highlighted by the different colours for \textit{No Prediction} (red), \textit{CVM with Prediction} (yellow), \textit{Proposed Method with Prediction} (green). We find that without accounting for the prediction trajectory of the person, the robot collides with the person. In contrast, our proposed framework can avoid collision and complete the task.}
\label{fig:change_places_comp}
\end{figure*}
To demonstrate the robustness and capabilities of our whole integrated framework, we deploy our implementation on a physical Toyota Human Support Robot and explore both base-only and whole-body tasks across a range of dynamic scenarios as follows:
\begin{enumerate}
    \item Change of Places
    \item Change of Places with Obstacle
    \item Multi-Goal
    \item Narrow Passage
    \item Change of Places with Half-Wall
\end{enumerate}

Note that in base-only tasks, the motion planner still optimises in the high-dimensional space of whole-body motions. The baselines of interest for this work are: (1) \textit{No Prediction} and (2) Prediction using \textit{CVM}.

In \citep{FineanIros2021}, the robot was able to adapt to changes in the environment; however, the human's trajectory was not aimed directly towards the robot in any of the tasks. Hence, the robot was able to perform sufficiently well without prediction. In contrast, for most tasks presented in this work, the robot is required to move out of the way of a person in order to avoid collision.

In our hardware experiments, we perform calculations on an external laptop connected to the HSR via an Ethernet connection to provide sufficient bandwidth for the transfer of images. Hardware specifications for the laptop are: NVIDIA RTX 2070 Super GPU, 8-core Intel Core i9-10980HK CPU @ \SI{5.30}{\giga\hertz} and \SI{2667}{\mega\hertz} DDR4 RAM.

\paragraph{Results}
Due to the relatively high speed of the human motions in our hardware experiments, we found that the robot always ended up in collision without accounting for the human's predicted motion. We provide supplementary video footage of these experiments\footnote{Supplementary video available at \url{https://youtu.be/gdC3mpZNjG4}} and describe the results of each experiment in the following subsections.

\subsection{\textit{Change of Places} - Prediction is Needed}
In our simplest task, \textit{Change of Places}, we do not provide static obstacles, and the robot is tasked with a base-only goal in front of a person. During the task, the person walks towards a goal behind the robot. The resultant task is illustrated in Fig.~\ref{fig:change_places_comp}. We observed that in the \textit{No Prediction} case, without predicting the human's trajectory, re-planned robot trajectories repeatedly become invalid, resulting in collision. Due to the straight-line nature of this task, we found that the \textit{CVM} performed equally as well as our full prediction method. For brevity, in further tasks, we only consider the \textit{CVM} baseline.

\subsection{\textit{Change of Places with Obstacle} - Limitations of the CVM}
\begin{figure}[htb]
        \centering
          \subfloat[\textit{Change of Places with Obstacle}\label{fig:change_places_obs}]{
          \includegraphics[draft=false,width=01.0\columnwidth]{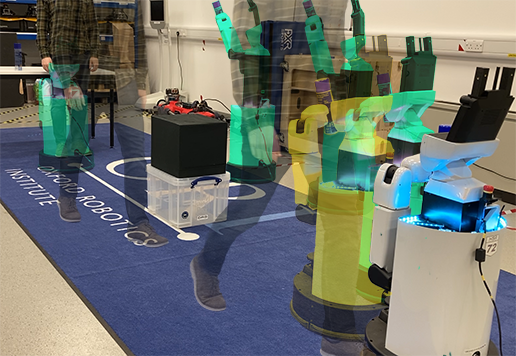}}   
          \vfill
          \subfloat[CVM Prediction\label{fig:change_places_obs_cvm}]{
          \includegraphics[draft=false,width=0.49\columnwidth]{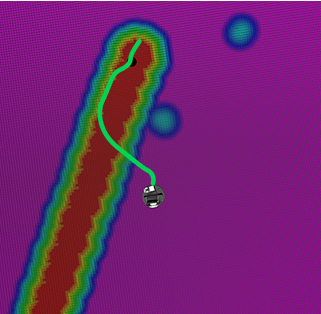}}  
        %   \hfill
          \subfloat[Our Prediction\label{fig:change_places_obs_ours}]{
          \includegraphics[draft=false,width=0.49\columnwidth]{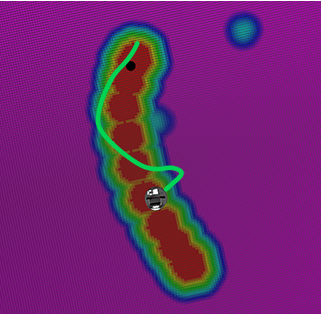}}  
\caption{\textit{Change of Places with Obstacle} -- a person walks around a static obstacle towards the robot's starting location. Meanwhile, the robot is tasked with a whole-body goal to place a canister on top of a table on the other side of the room. Fig.~\ref{fig:change_places_obs} shows the trajectories taken by our method (green) and using the \textit{CVM} (yellow). While our methods avoided the person, the \textit{CVM} trajectory did not move out of the way in time, requiring the human to slow down to avoid a collision. The reason for this is explained by Figures \ref{fig:change_places_obs_cvm} and \ref{fig:change_places_obs_ours} which show superposed 2D projections of the 3D composite distance fields used for motion planning at two given moments in time. While our prediction method more accurately predicts the human's trajectory, the \textit{CVM} predicts a trajectory tangential to the actual one taken.}
\label{fig:change_places_obs_comp}
\end{figure}

With the addition of a central obstacle to the previous task, both the robot and human must take curved geodesics. Results are illustrated in Fig.~\ref{fig:change_places_obs_comp}. This experiment highlighted the limitation of the \textit{CVM} -- when a person follows a curved path, the resultant prediction is tangential to the actual path. In this task, this erroneous prediction results in collision. In contrast, our method accurately predicted the person's trajectory, enabling the robot to follow a collision-free trajectory. 

\subsection{\textit{Multi-Goal} - Robust to Intention Recognition}
\begin{figure}[htb]
\centering
  \subfloat[\label{fig:multigoal1}]{
  \includegraphics[draft=false, width=1.0\columnwidth]{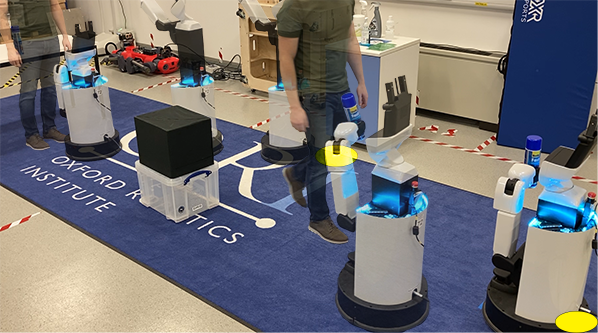}}   
  \vfill
  \subfloat[\label{fig:multigoal2}]{
  \includegraphics[draft=false,width=0.32\columnwidth]{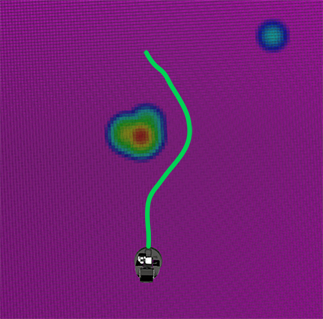}}
    \subfloat[\label{fig:multigoal3}]{
  \includegraphics[draft=false,width=0.32\columnwidth]{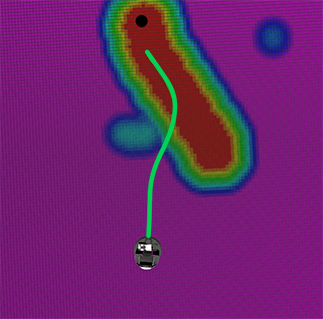}}   
  \subfloat[\label{fig:multigoal4}]{
  \includegraphics[draft=false,width=0.32\columnwidth]{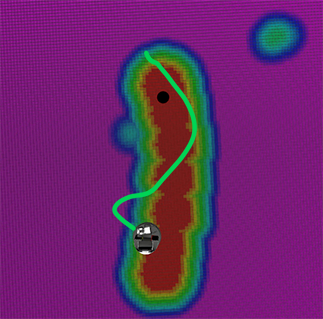}}   
\caption{\textit{Multi-Goal} -- the robot is tasked with a whole-body goal to place a canister on a table on the other side of a static obstacle. A person in the scene has two possible goal locations - the first is behind the robot's starting location, the second is at a hand-wash station in front of the robot. This experiment demonstrated our framework's ability to adapt and update human trajectory predictions even when the human's intended goal is deemed to have changed. Figure~\ref{fig:multigoal2} shows the initial planned robot trajectory superposed on an aerial view of the ground distance field. Figure~\ref{fig:multigoal3} shows the updated trajectory as the human is deemed to be moving towards the hand-wash station while Fig.~\ref{fig:multigoal3} shows the updated trajectory as the prediction module correctly identifies that the person's intended goal is the one behind the robot.}
\label{fig:multi-goal}
\end{figure}

While the goal-based aspect of our prediction framework is more extensively evaluated in Sec.~\ref{sec:prediction_eval}, we provide a hardware experiment in which the person is determined to have two potential goals: one behind the robot's starting position, the other at a hand-wash station across the robot's path. The robot is tasked with a whole-body motion to place a can on the table opposite. Across multiple trials, the robot successfully predicts the person's intended goal and adapts its motions appropriately to execute the task collision-free. Figure~\ref{fig:multi-goal} illustrates these results.

\subsection{Additional Capabilities}
\begin{figure}[htb]
        \centering
          \includegraphics[draft=false,width=1.0\columnwidth]{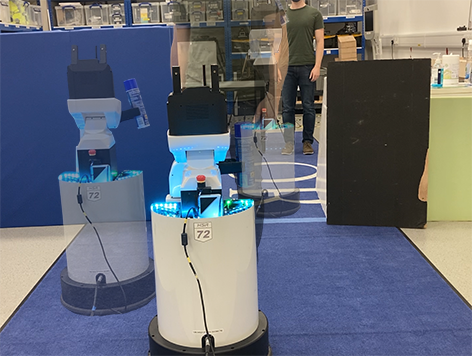}
\caption{\textit{Narrow Passage} -- the robot is tasked with a whole-body goal to place a canister on top of a table that is on the other side of a narrow passage. During execution, a person walks through the narrow passage to act as a dynamic obstacle. Our method achieved a successful collision-free trajectory by re-planning to move to the side while the person walked past.}
\label{fig:narrow_passage}
\end{figure}

We additionally tested our approach in a \textit{Narrow Passage} task and a variant of the \textit{Change of Places with Obstacle} experiment in which the central obstacle was replaced with a wall across half of the room. In both tasks, the robot successfully adapted to the predicted trajectory of the person and moved to the side of the person's path before continuing towards the goal. Images from the two tasks are shown in Figures \ref{fig:front_page} and \ref{fig:narrow_passage}.

\section{Discussion}
One advantage of our proposed human trajectory prediction approach is that it bridges the gap between model-based and learning-based prediction methods. A major limitation of learnt models is their ability to transfer to scenarios that differ from those in which it was trained. In contrast, an appealing attribute of our human trajectory prediction method is that it can be readily deployed in any environment by defaulting to a constant-velocity model while retaining the ability to improve over time as a prior distribution is learnt over likely human goal intentions. An interesting direction for further research would be to explore the online-learning of human intentions further and incorporate scene semantics.

In \citep{FineanIros2021}, no additional filtering was applied to the maintained voxelmap to remove `lingering' voxels that a moving obstacle may leave. In this work, we found that these lingering voxels provided a substantial disadvantage for motion planning compared to our proposed method; the proposed method uses dilated segmentation of dynamic obstacles, resulting in a cleaner static voxelmap. As such, to provide an appropriate \textit{No Prediction} baseline in line with our previous work, we introduced the segmentation pipeline such that the human obstacle is tracked and composited into the singular distance field used for the motion planning.

While we obtained robust re-planning and collision avoidance behaviours across various tasks, there are several limitations in the presented work that are worth noting. Firstly, we do not explicitly model uncertainty in our current method of compositing the predicted positions of moving obstacles.
Instead, we account for a margin-of-safety via the $\epsilon$ parameter within the GPMP2-based obstacle factors -- $\epsilon$ determines the upper distance used for hinge-loss obstacle costs. While further exploration of this was beyond the scope of the presented work, we could enlarge the volume of the person/cylinder over the course of the prediction time horizon to appropriately account for a growing uncertainty in future position as time increases. 

In this work, we demonstrated that GPU implementations of predicted composite distance fields can provide a significant performance boost compared to calculating distance fields from scratch. However, as shown in Fig.~\ref{fig:composite_sdf_benchmarking}, the key bottleneck for further composite distance field performance gains is the device-host transfer time. Future work could explore alternative approaches to minimise data transfer between device and host.

A natural limitation of our motion planning implementation is that the optimisation is prone to get stuck in local minima by only optimising a single trajectory. While this did not result in collisions in our experiments, the phenomenon is evident in trajectories such as Figures \ref{fig:change_places_obs_ours} and \ref{fig:multigoal3} -- rather than planning to travel on the other side of the static obstacle to the human, the re-planned trajectory avoids the human but stays within the same homotopy class. To address this, one could consider maintaining and optimising multiple trajectories at a time in different homotopy classes, such as work by \cite{Kolur2019} and \cite{merkt2021memoryclustering}, however, this is likely to increase the planning time and limit the robot's ability to react.

It is worth noting that we use two different motion planning approaches in our proposed framework. For robot motion planning, we use predicted versions of the environment for each time step, while for human trajectory prediction, we only use the latest observation of the environment. Our reasoning for this is two-fold; firstly, the walking speed of a human is significantly higher than that of the robot's mobile speed, so a human has less need to account for the predicted trajectory of the robot. Secondly, from our experience, humans will readily travel much closer towards the path of a moving robot, while from the robot perspective, we need to retain a more cautious approach to collision avoidance.

\section{Conclusion}
To enable predictive whole-body motion planning in dynamic environments, we introduced several novel methods and integrated them within a novel framework that can account for the predicted trajectories of humans in a scene. We firstly proposed an intention-aware trajectory prediction model for humans in indoor environments and demonstrated state-of-the-art performance on both a publicly available dataset as well as our own goal-oriented dataset, the Oxford Indoor Human Motion (Oxford-IHM) dataset, that we make publicly available. 

For predictive and reactive motion planning, we proposed the Receding Horizon And Predictive Gaussian Process Motion Planner~2 (RHAP-GPMP2), a receding-horizon motion planner that utilising predicted composite distance fields to embed the predicted trajectories of moving obstacles. To this end, we demonstrated the viability and effectiveness of composite distance fields in a GPU-based perception framework and show that composite distance fields can reduce distance field computation times by \SIrange{89}{93}{\percent}, underpinning our integrated framework's ability to avoid moving obstacles in real-world environments.

We verified our proposed framework on a physical Toyota Human Support Robot (HSR) and demonstrated that our system can use live sensor measurements to predict and incorporate the trajectories of humans in a robot's workspace, enabling it to avoid collisions when performing whole-body motion planning across a variety of challenging and dynamic environments.

\bibliographystyle{SageH}
\bibliography{main}

\begin{thebibliography}{76}
\providecommand{\natexlab}[1]{#1}
\providecommand{\url}[1]{\texttt{#1}}
\providecommand{\urlprefix}{URL }
\expandafter\ifx\csname urlstyle\endcsname\relax
  \providecommand{\doi}[1]{DOI:\discretionary{}{}{}#1}\else
  \providecommand{\doi}{DOI:\discretionary{}{}{}\begingroup
  \urlstyle{rm}\Url}\fi

\bibitem[{Alahi et~al.(2016)Alahi, Goel, Ramanathan, Robicquet, Fei-Fei and
  Savarese}]{alahi2016social}
Alahi A, Goel K, Ramanathan V, Robicquet A, Fei-Fei L and Savarese S (2016)
  Social lstm: Human trajectory prediction in crowded spaces.
\newblock In: \emph{Proceedings of the IEEE conference on computer vision and
  pattern recognition}. pp. 961--971.

\bibitem[{Amirian et~al.(2019)Amirian, Hayet and
  Pettr{\'e}}]{amirian2019social}
Amirian J, Hayet JB and Pettr{\'e} J (2019) Social ways: Learning multi-modal
  distributions of pedestrian trajectories with gans.
\newblock In: \emph{Proceedings of the IEEE/CVF Conference on Computer Vision
  and Pattern Recognition Workshops}. pp. 0--0.

\bibitem[{Bartoli et~al.(2018)Bartoli, Lisanti, Ballan and
  Del~Bimbo}]{bartoli2018context}
Bartoli F, Lisanti G, Ballan L and Del~Bimbo A (2018) Context-aware trajectory
  prediction.
\newblock In: \emph{2018 24th International Conference on Pattern Recognition
  (ICPR)}. IEEE, pp. 1941--1946.

\bibitem[{Batkovic et~al.(2018)Batkovic, Zanon, Lubbe and
  Falcone}]{Batkovic2018}
Batkovic I, Zanon M, Lubbe N and Falcone P (2018) A computationally efficient
  model for pedestrian motion prediction.
\newblock In: \emph{2018 European Control Conference (ECC)}. IEEE.
\newblock ISBN 978-3-9524-2698-2, pp. 374--379.
\newblock \doi{10.23919/ECC.2018.8550300}.

\bibitem[{Bernardin and Stiefelhagen(2008)}]{Bernardin2008}
Bernardin K and Stiefelhagen R (2008) Evaluating multiple object tracking
  performance: The clear mot metrics.
\newblock \emph{EURASIP Journal on Image and Video Processing}
  \doi{10.1155/2008/246309}.

\bibitem[{Best and Fitch(2015)}]{best2015bayesian}
Best G and Fitch R (2015) Bayesian intention inference for trajectory
  prediction with an unknown goal destination.
\newblock In: \emph{2015 IEEE/RSJ International Conference on Intelligent
  Robots and Systems (IROS)}. IEEE, pp. 5817--5823.

\bibitem[{Brscic et~al.(2013)Brscic, Kanda, Ikeda and Miyashita}]{Brscic2013}
Brscic D, Kanda T, Ikeda T and Miyashita T (2013) Person tracking in large
  public spaces using 3-d range sensors.
\newblock \emph{IEEE Transactions on Human-Machine Systems} 43: 522--534.
\newblock \doi{10.1109/THMS.2013.2283945}.

\bibitem[{Cao et~al.(2010)Cao, Tang, Mohamed and Tan}]{PBA}
Cao TT, Tang K, Mohamed A and Tan TS (2010) {Parallel Banding Algorithm to
  compute exact distance transform with the GPU}.
\newblock In: \emph{ACM SIGGRAPH I3D}.
\newblock ISBN 9781605589381, pp. 83--90.
\newblock \doi{10.1145/1730804.1730818}.

\bibitem[{Cao et~al.(2020)Cao, Gao, Mangalam, Cai, Vo and Malik}]{cao2020long}
Cao Z, Gao H, Mangalam K, Cai QZ, Vo M and Malik J (2020) Long-term human
  motion prediction with scene context.
\newblock In: \emph{European Conference on Computer Vision}. Springer, pp.
  387--404.

\bibitem[{Cao et~al.(2021)Cao, Hidalgo, Simon, Wei and Sheikh}]{Cao2021}
Cao Z, Hidalgo G, Simon T, Wei SE and Sheikh Y (2021) {OpenPose: Realtime
  Multi-Person 2D Pose Estimation Using Part Affinity Fields}.
\newblock \emph{IEEE Transactions on Pattern Analysis and Machine Intelligence}
  43(1): 172--186.
\newblock \doi{10.1109/TPAMI.2019.2929257}.

\bibitem[{Dellaert(2012)}]{dellaert2012factor}
Dellaert F (2012) Factor graphs and gtsam: A hands-on introduction.
\newblock Technical report, Georgia Institute of Technology.

\bibitem[{Dou et~al.(2015)Dou, Taylor, Fuchs, Fitzgibbon and Izadi}]{Dou2015}
Dou M, Taylor J, Fuchs H, Fitzgibbon A and Izadi S (2015) 3d scanning
  deformable objects with a single rgbd sensor.
\newblock IEEE.
\newblock ISBN 978-1-4673-6964-0, pp. 493--501.
\newblock \doi{10.1109/CVPR.2015.7298647}.
\newblock \urlprefix\url{http://ieeexplore.ieee.org/document/7298647/}.

\bibitem[{Finean et~al.(2020)Finean, Merkt and Havoutis}]{FineanComposite2020}
Finean MN, Merkt W and Havoutis I (2020) {Predicted composite signed-distance
  fields for real-time motion planning in dynamic environments}.
\newblock In: \emph{International Conference Conf. Automat. Planning and
  Scheduling}.

\bibitem[{Finean et~al.(2021)Finean, Merkt and Havoutis}]{FineanIros2021}
Finean MN, Merkt W and Havoutis I (2021) {Simultaneous Scene Reconstruction and
  Whole-Body Motion Planning for Safe Operation in Dynamic Environments}.
\newblock In: \emph{IEEE/RSJ International Conference on Intelligent Robots and
  Systems}.

\bibitem[{Flores et~al.(2019)Flores, Merdrignac, de~Charette, Navas, Milanes
  and Nashashibi}]{Flores2019}
Flores C, Merdrignac P, de~Charette R, Navas F, Milanes V and Nashashibi F
  (2019) A cooperative car-following/emergency braking system with
  prediction-based pedestrian avoidance capabilities.
\newblock \emph{IEEE Transactions on Intelligent Transportation Systems} 20:
  1837--1846.
\newblock \doi{10.1109/TITS.2018.2841644}.

\bibitem[{Ghosh et~al.(2017)Ghosh, Song, Aksan and
  Hilliges}]{ghosh2017learning}
Ghosh P, Song J, Aksan E and Hilliges O (2017) Learning human motion models for
  long-term predictions.
\newblock In: \emph{2017 International Conference on 3D Vision (3DV)}. IEEE,
  pp. 458--466.

\bibitem[{Han et~al.(2019)Han, Gao, Zhou and Shen}]{FIESTA}
Han L, Gao F, Zhou B and Shen S (2019) {FIESTA: Fast Incremental Euclidean
  Distance Fields for Online Motion Planning of Aerial Robots}.
\newblock In: \emph{IEEE/RSJ International Conference on Intelligent Robots and
  Systems}.
\newblock ISBN 9781728140049, pp. 4423--4430.
\newblock \doi{10.1109/IROS40897.2019.8968199}.

\bibitem[{Handa et~al.(2014)Handa, Whelan, McDonald and Davison}]{Handa2014}
Handa A, Whelan T, McDonald J and Davison AJ (2014) A benchmark for rgb-d
  visual odometry, 3d reconstruction and slam.
\newblock In: \emph{IEEE International Conference on Robotics and Automation}.
  IEEE.
\newblock ISBN 978-1-4799-3685-4, pp. 1524--1531.
\newblock \doi{10.1109/ICRA.2014.6907054}.

\bibitem[{He et~al.(2020)He, Gkioxari, Doll{\'{a}}r and Girshick}]{He2020}
He K, Gkioxari G, Doll{\'{a}}r P and Girshick R (2020) {Mask R-CNN}.
\newblock \emph{IEEE Transactions on Pattern Analysis and Machine Intelligence}
  42(2): 386--397.
\newblock \doi{10.1109/TPAMI.2018.2844175}.

\bibitem[{Helbing and Molnar(1995)}]{helbing1995social}
Helbing D and Molnar P (1995) Social force model for pedestrian dynamics.
\newblock \emph{Physical review E} 51(5): 4282.

\bibitem[{Hermann et~al.(2014)Hermann, Drews, Bauer, Klemm, Roennau and
  Dillmann}]{GPUVoxels}
Hermann A, Drews F, Bauer J, Klemm S, Roennau A and Dillmann R (2014) {Unified
  GPU voxel collision detection for mobile manipulation planning}.
\newblock In: \emph{IEEE/RSJ International Conference on Intelligent Robots and
  Systems}.
\newblock ISBN 9781479969340, pp. 4154--4160.
\newblock \doi{10.1109/IROS.2014.6943148}.

\bibitem[{Hermann et~al.(2015)Hermann, Mauch, Fischnaller, Klemm, Roennau and
  Dillmann}]{hermann2015anticipate}
Hermann A, Mauch F, Fischnaller K, Klemm S, Roennau A and Dillmann R (2015)
  Anticipate your surroundings: Predictive collision detection between dynamic
  obstacles and planned robot trajectories on the gpu.
\newblock In: \emph{2015 European Conference on Mobile Robots (ECMR)}. IEEE,
  pp. 1--8.

\bibitem[{Janoch et~al.(2011)Janoch, Karayev, Jia, Barron, Fritz, Saenko and
  Darrell}]{Janoch2011}
Janoch A, Karayev S, Jia Y, Barron JT, Fritz M, Saenko K and Darrell T (2011) A
  category-level 3-d object dataset: Putting the kinect to work.
\newblock In: \emph{IEEE International Conference on Computer Vision Workshops
  (ICCV Workshops)}. IEEE.
\newblock ISBN 978-1-4673-0063-6, pp. 1168--1174.
\newblock \doi{10.1109/ICCVW.2011.6130382}.

\bibitem[{Juelg et~al.(2018)Juelg, Hermann, Roennau and
  Dillmann}]{GPUVoxelsMobile}
Juelg C, Hermann A, Roennau A and Dillmann R (2018) {Fast online collision
  avoidance for mobile service robots through potential fields on 3D
  environment data processed on GPUs}.
\newblock In: \emph{IEEE International Conference on Robotics and Biomimetics
  (ROBIO)}.
\newblock ISBN 9781538637418.
\newblock \doi{10.1109/ROBIO.2017.8324535}.

\bibitem[{Kalakrishnan et~al.(2011)Kalakrishnan, Chitta, Theodorou, Pastor and
  Schaal}]{Kalakrishnan2011}
Kalakrishnan M, Chitta S, Theodorou E, Pastor P and Schaal S (2011) Stomp:
  Stochastic trajectory optimization for motion planning.
\newblock In: \emph{IEEE International Conference on Robotics and Automation}.
  IEEE.
\newblock ISBN 978-1-61284-386-5, pp. 4569--4574.
\newblock \doi{10.1109/ICRA.2011.5980280}.

\bibitem[{Karamouzas et~al.(2009)Karamouzas, Heil, Van~Beek and
  Overmars}]{karamouzas2009predictive}
Karamouzas I, Heil P, Van~Beek P and Overmars MH (2009) A predictive collision
  avoidance model for pedestrian simulation.
\newblock In: \emph{International workshop on motion in games}. Springer, pp.
  41--52.

\bibitem[{Kellnhofer et~al.(2019)Kellnhofer, Recasens, Stent, Matusik and
  Torralba}]{kellnhofer2019gaze360}
Kellnhofer P, Recasens A, Stent S, Matusik W and Torralba A (2019) Gaze360:
  Physically unconstrained gaze estimation in the wild.
\newblock In: \emph{Proceedings of the IEEE/CVF International Conference on
  Computer Vision}. pp. 6912--6921.

\bibitem[{Kitani et~al.(2012)Kitani, Ziebart, Bagnell and
  Hebert}]{kitani2012activity}
Kitani KM, Ziebart BD, Bagnell JA and Hebert M (2012) Activity forecasting.
\newblock In: \emph{European conference on computer vision}. Springer, pp.
  201--214.

\bibitem[{Kohlbrecher et~al.(2011)Kohlbrecher, Meyer, von Stryk and
  Klingauf}]{hectormapping}
Kohlbrecher S, Meyer J, von Stryk O and Klingauf U (2011) A flexible and
  scalable slam system with full 3d motion estimation.
\newblock In: \emph{Proc. IEEE International Symposium on Safety, Security and
  Rescue Robotics (SSRR)}. IEEE.

\bibitem[{Kolur et~al.(2019)Kolur, Chintalapudi, Boots and Mukadam}]{Kolur2019}
Kolur K, Chintalapudi S, Boots B and Mukadam M (2019) Online motion planning
  over multiple homotopy classes with gaussian process inference.
\newblock In: \emph{IEEE/RSJ International Conference on Intelligent Robots and
  Systems}.
\newblock ISBN 978-1-7281-4004-9, pp. 2358--2364.
\newblock \doi{10.1109/IROS40897.2019.8967598}.

\bibitem[{Kratzer et~al.(2021)Kratzer, Bihlmaier, Midlagajni, Prakash,
  Toussaint and Mainprice}]{Kratzer2021}
Kratzer P, Bihlmaier S, Midlagajni NB, Prakash R, Toussaint M and Mainprice J
  (2021) Mogaze: A dataset of full-body motions that includes workspace
  geometry and eye-gaze.
\newblock \emph{IEEE Robotics and Automation Letters} 6: 367--373.
\newblock \doi{10.1109/LRA.2020.3043167}.
\newblock \urlprefix\url{https://ieeexplore.ieee.org/document/9286421/}.

\bibitem[{Kratzer et~al.(2020)Kratzer, Toussaint and
  Mainprice}]{kratzer2020prediction}
Kratzer P, Toussaint M and Mainprice J (2020) Prediction of human full-body
  movements with motion optimization and recurrent neural networks.
\newblock In: \emph{2020 IEEE International Conference on Robotics and
  Automation (ICRA)}. IEEE, pp. 1792--1798.

\bibitem[{Kschischang et~al.(2001)Kschischang, Frey and
  Loeliger}]{kschischang2001factor}
Kschischang FR, Frey BJ and Loeliger HA (2001) Factor graphs and the
  sum-product algorithm.
\newblock \emph{IEEE Transactions on Information Theory} 47(2): 498--519.

\bibitem[{Lai et~al.(2011)Lai, Bo, Ren and Fox}]{Lai2011}
Lai K, Bo L, Ren X and Fox D (2011) A large-scale hierarchical multi-view rgb-d
  object dataset.
\newblock In: \emph{IEEE International Conference on Robotics and Automation}.
  IEEE.
\newblock ISBN 978-1-61284-386-5, pp. 1817--1824.
\newblock \doi{10.1109/ICRA.2011.5980382}.

\bibitem[{Lee and Park(2020)}]{Lee2020}
Lee Y and Park J (2020) {CenterMask: Real-time anchor-free instance
  segmentation}.
\newblock In: \emph{Proceedings of the IEEE Computer Society Conference on
  Computer Vision and Pattern Recognition}.
\newblock ISBN 1911.06667v6, pp. 13903--13912.
\newblock \doi{10.1109/CVPR42600.2020.01392}.

\bibitem[{Luo et~al.(2018)Luo, Cai, Bera, Hsu, Lee and Manocha}]{Luo2018}
Luo Y, Cai P, Bera A, Hsu D, Lee WS and Manocha D (2018) Porca: Modeling and
  planning for autonomous driving among many pedestrians.
\newblock \emph{IEEE Robotics and Automation Letters} 3: 3418--3425.
\newblock \doi{10.1109/LRA.2018.2852793}.

\bibitem[{Mainprice and Berenson(2013)}]{Mainprice2013}
Mainprice J and Berenson D (2013) {Human-robot collaborative manipulation
  planning using early prediction of human motion}.
\newblock In: \emph{IEEE/RSJ International Conference on Intelligent Robots and
  Systems}.
\newblock ISBN 9781467363587, pp. 299--306.
\newblock \doi{10.1109/IROS.2013.6696368}.

\bibitem[{Merkt et~al.(2021)Merkt, Ivan, Dinev, Havoutis and
  Vijayakumar}]{merkt2021memoryclustering}
Merkt WX, Ivan V, Dinev T, Havoutis I and Vijayakumar S (2021) Memory
  clustering using persistent homology for multimodality- and
  discontinuity-sensitive learning of optimal control warm-starts.
\newblock \emph{IEEE Transactions on Robotics} 37(5): 1649--1660.
\newblock \doi{10.1109/TRO.2021.3069132}.

\bibitem[{Mukadam et~al.(2018)Mukadam, Dong, Yan, Dellaert and Boots}]{gpmp2}
Mukadam M, Dong J, Yan X, Dellaert F and Boots B (2018) {Continuous-time
  Gaussian process motion planning via probabilistic inference}.
\newblock \emph{The Int. J. of Rob. Res.} 37(11): 1319--1340.
\newblock \doi{10.1177/0278364918790369}.

\bibitem[{Munaro and Menegatti(2014)}]{Munaro2014}
Munaro M and Menegatti E (2014) {Fast RGB-D people tracking for service
  robots}.
\newblock \emph{Autonomous Robots} 37(3): 227--242.
\newblock \doi{10.1007/s10514-014-9385-0}.

\bibitem[{Newcombe et~al.(2011)Newcombe, Fitzgibbon, Izadi, Hilliges,
  Molyneaux, Kim, Davison, Kohi, Shotton and Hodges}]{Newcombe2011}
Newcombe RA, Fitzgibbon A, Izadi S, Hilliges O, Molyneaux D, Kim D, Davison AJ,
  Kohi P, Shotton J and Hodges S (2011) {KinectFusion: Real-time dense surface
  mapping and tracking}.
\newblock In: \emph{2011 10th IEEE International Symposium on Mixed and
  Augmented Reality}. IEEE.
\newblock ISBN 978-1-4577-2183-0, pp. 127--136.
\newblock \doi{10.1109/ISMAR.2011.6092378}.

\bibitem[{Oleynikova et~al.(2016)Oleynikova, Burri, Taylor, Nieto, Siegwart and
  Galceran}]{Oleynikova2016replanning}
Oleynikova H, Burri M, Taylor Z, Nieto J, Siegwart R and Galceran E (2016)
  Continuous-time trajectory optimization for online uav replanning.
\newblock In: \emph{IEEE/RSJ International Conference on Intelligent Robots and
  Systems}, volume 2016-November. IEEE.
\newblock ISBN 978-1-5090-3762-9, pp. 5332--5339.
\newblock \doi{10.1109/IROS.2016.7759784}.

\bibitem[{Oleynikova et~al.(2017)Oleynikova, Taylor, Fehr, Siegwart and
  Nieto}]{Oleynikova2017}
Oleynikova H, Taylor Z, Fehr M, Siegwart R and Nieto J (2017) {Voxblox:
  Incremental 3D Euclidean Signed Distance Fields for on-board MAV planning}.
\newblock In: \emph{IEEE/RSJ International Conference on Intelligent Robots and
  Systems}, volume 2017-September.
\newblock ISBN 9781538626825, pp. 1366--1373.
\newblock \doi{10.1109/IROS.2017.8202315}.

\bibitem[{Palazzolo et~al.(2019)Palazzolo, Behley, Lottes, Giguere and
  Stachniss}]{Palazzolo2019}
Palazzolo E, Behley J, Lottes P, Giguere P and Stachniss C (2019) {ReFusion: 3D
  Reconstruction in Dynamic Environments for RGB-D Cameras Exploiting
  Residuals}.
\newblock In: \emph{IEEE/RSJ International Conference on Intelligent Robots and
  Systems}.
\newblock ISBN 9781728140049, pp. 7855--7862.
\newblock \doi{10.1109/IROS40897.2019.8967590}.

\bibitem[{Park et~al.(2012)Park, Pan and Manocha}]{Park2012}
Park C, Pan J and Manocha D (2012) {ITOMP: Incremental trajectory optimization
  for real-time replanning in dynamic environments}.
\newblock In: \emph{International Conference Conf. Automat. Planning and
  Scheduling}.
\newblock ISBN 9781577355625, pp. 207--215.

\bibitem[{Park et~al.(2019)Park, Park and Manocha}]{Park2019}
Park JS, Park C and Manocha D (2019) {I-Planner: Intention-aware motion
  planning using learning-based human motion prediction}.
\newblock \emph{The Int. J. of Rob. Res.} 38(1): 23--39.
\newblock \doi{10.1177/0278364918812981}.

\bibitem[{Park et~al.(2018)Park, Spurr and Hilliges}]{park2018deep}
Park S, Spurr A and Hilliges O (2018) Deep pictorial gaze estimation.
\newblock In: \emph{Proceedings of the European Conference on Computer Vision
  (ECCV)}. pp. 721--738.

\bibitem[{Pellegrini et~al.(2009)Pellegrini, Ess, Schindler and
  Van~Gool}]{pellegrini2009you}
Pellegrini S, Ess A, Schindler K and Van~Gool L (2009) You'll never walk alone:
  Modeling social behavior for multi-target tracking.
\newblock In: \emph{2009 IEEE 12th International Conference on Computer
  Vision}. IEEE, pp. 261--268.

\bibitem[{Rehder et~al.(2018)Rehder, Wirth, Lauer and Stiller}]{Rehder2018}
Rehder E, Wirth F, Lauer M and Stiller C (2018) Pedestrian prediction by
  planning using deep neural networks.
\newblock In: \emph{IEEE International Conference on Robotics and Automation}.
  IEEE.
\newblock ISBN 978-1-5386-3081-5, pp. 1--5.
\newblock \doi{10.1109/ICRA.2018.8460203}.

\bibitem[{Ren et~al.(2017)Ren, He, Girshick and Sun}]{Ren2017}
Ren S, He K, Girshick R and Sun J (2017) Faster r-cnn: Towards real-time object
  detection with region proposal networks.
\newblock \emph{IEEE Transactions on Pattern Analysis and Machine Intelligence}
  39: 1137--1149.
\newblock \doi{10.1109/TPAMI.2016.2577031}.
\newblock \urlprefix\url{http://ieeexplore.ieee.org/document/7485869/}.

\bibitem[{R{\"o}smann et~al.(2017)R{\"o}smann, Oeljeklaus, Hoffmann and
  Bertram}]{rosmann2017online}
R{\"o}smann C, Oeljeklaus M, Hoffmann F and Bertram T (2017) Online trajectory
  prediction and planning for social robot navigation.
\newblock In: \emph{2017 IEEE International Conference on Advanced Intelligent
  Mechatronics (AIM)}. IEEE, pp. 1255--1260.

\bibitem[{Rudenko et~al.(2021)Rudenko, Huang, Palmieri, Arras and
  Lilienthal}]{Rudenko2021}
Rudenko A, Huang W, Palmieri L, Arras KO and Lilienthal AJ (2021) {Atlas : a
  Benchmarking Tool for Human Motion Prediction Algorithms}.
\newblock \emph{Robotics: Science and Systems (RSS) Workshop on Social Robot
  Navigation} .

\bibitem[{Rudenko et~al.(2020{\natexlab{a}})Rudenko, Kucner, Swaminathan,
  Chadalavada, Arras and Lilienthal}]{thorDataset2019}
Rudenko A, Kucner TP, Swaminathan CS, Chadalavada RT, Arras KO and Lilienthal
  AJ (2020{\natexlab{a}}) Th{\"o}r: Human-robot navigation data collection and
  accurate motion trajectories dataset.
\newblock \emph{IEEE Robotics and Automation Letters} 5(2): 676--682.

\bibitem[{Rudenko et~al.(2018)Rudenko, Palmieri and Arras}]{rudenko2018joint}
Rudenko A, Palmieri L and Arras KO (2018) Joint long-term prediction of human
  motion using a planning-based social force approach.
\newblock In: \emph{2018 IEEE International Conference on Robotics and
  Automation (ICRA)}. IEEE, pp. 4571--4577.

\bibitem[{Rudenko et~al.(2020{\natexlab{b}})Rudenko, Palmieri, Herman, Kitani,
  Gavrila and Arras}]{rudenko2020human}
Rudenko A, Palmieri L, Herman M, Kitani KM, Gavrila DM and Arras KO
  (2020{\natexlab{b}}) Human motion trajectory prediction: A survey.
\newblock \emph{The Int. J. of Rob. Res.} 39(8): 895--935.

\bibitem[{Runz et~al.(2019)Runz, Buffier and Agapito}]{Runz2019}
Runz M, Buffier M and Agapito L (2019) {MaskFusion: Real-Time Recognition,
  Tracking and Reconstruction of Multiple Moving Objects}.
\newblock In: \emph{Proceedings of the 2018 IEEE International Symposium on
  Mixed and Augmented Reality, ISMAR 2018}.
\newblock ISBN 9781538674598, pp. 10--20.
\newblock \doi{10.1109/ISMAR.2018.00024}.

\bibitem[{Sch{\"o}ller et~al.(2020)Sch{\"o}ller, Aravantinos, Lay and
  Knoll}]{scholler2020constant}
Sch{\"o}ller C, Aravantinos V, Lay F and Knoll A (2020) What the constant
  velocity model can teach us about pedestrian motion prediction.
\newblock \emph{IEEE Robotics and Automation Letters} 5(2): 1696--1703.

\bibitem[{Scona et~al.(2018)Scona, Jaimez, Petillot, Fallon and
  Cremers}]{Scona2018}
Scona R, Jaimez M, Petillot YR, Fallon M and Cremers D (2018) {StaticFusion:
  Background Reconstruction for Dense RGB-D SLAM in Dynamic Environments}.
\newblock In: \emph{IEEE International Conference on Robotics and Automation}.
\newblock ISBN 9781538630815, pp. 3849--3856.
\newblock \doi{10.1109/ICRA.2018.8460681}.

\bibitem[{Sturm et~al.(2012)Sturm, Engelhard, Endres, Burgard and
  Cremers}]{Sturm2012}
Sturm J, Engelhard N, Endres F, Burgard W and Cremers D (2012) {A benchmark for
  the evaluation of RGB-D SLAM systems}.
\newblock In: \emph{IEEE/RSJ International Conference on Intelligent Robots and
  Systems}.
\newblock ISBN 9781467317375, pp. 573--580.
\newblock \doi{10.1109/IROS.2012.6385773}.

\bibitem[{Sung et~al.(2012)Sung, Ponce, Selman and Saxena}]{Sung2012}
Sung J, Ponce C, Selman B and Saxena A (2012) Unstructured human activity
  detection from rgbd images.
\newblock In: \emph{IEEE International Conference on Robotics and Automation}.
  IEEE.
\newblock ISBN 978-1-4673-1405-3, pp. 842--849.
\newblock \doi{10.1109/ICRA.2012.6224591}.

\bibitem[{Treuille et~al.(2006)Treuille, Cooper and
  Popovi{\'c}}]{treuille2006continuum}
Treuille A, Cooper S and Popovi{\'c} Z (2006) Continuum crowds.
\newblock \emph{ACM Transactions on Graphics (TOG)} 25(3): 1160--1168.

\bibitem[{Vasquez(2016)}]{vasquez2016novel}
Vasquez D (2016) Novel planning-based algorithms for human motion prediction.
\newblock In: \emph{2016 IEEE International Conference on Robotics and
  Automation (ICRA)}. IEEE, pp. 3317--3322.

\bibitem[{Whelan et~al.(2012)Whelan, Kaess and Fallon}]{Whelan2012}
Whelan T, Kaess M and Fallon M (2012) {Kintinuous: Spatially extended
  kinectfusion}.
\newblock \emph{RSS Workshop on RGB-D: Advanced Reasoning with Depth Cameras} :
  7.

\bibitem[{Whelan et~al.(2015)Whelan, Leutenegger, Salas-Moreno, Glocker and
  Davison}]{Whelan2015}
Whelan T, Leutenegger S, Salas-Moreno RF, Glocker B and Davison AJ (2015)
  {ElasticFusion: Dense SLAM without a pose graph}.
\newblock In: \emph{Robotics: Science and Systems}, volume~11.
\newblock ISBN 9780992374716.
\newblock \doi{10.15607/RSS.2015.XI.001}.

\bibitem[{Xie et~al.(2021)Xie, Liu and Zheng}]{Xie2021}
Xie W, Liu PX and Zheng M (2021) {Moving Object Segmentation and Detection for
  Robust RGBD-SLAM in Dynamic Environments}.
\newblock \emph{IEEE Transactions on Instrumentation and Measurement} 70.
\newblock \doi{10.1109/TIM.2020.3026803}.

\bibitem[{Yan et~al.(2017)Yan, Duckett and Bellotto}]{yan2017online}
Yan Z, Duckett T and Bellotto N (2017) Online learning for human classification
  in 3d lidar-based tracking.
\newblock In: \emph{2017 IEEE/RSJ International Conference on Intelligent
  Robots and Systems (IROS)}. IEEE, pp. 864--871.

\bibitem[{Yu et~al.(2020)Yu, Ma, Ren, Zhao and Yi}]{yu2020spatio}
Yu C, Ma X, Ren J, Zhao H and Yi S (2020) Spatio-temporal graph transformer
  networks for pedestrian trajectory prediction.
\newblock In: \emph{European Conference on Computer Vision}. Springer, pp.
  507--523.

\bibitem[{Yu et~al.(2015)Yu, Russell, Campbell and Agapito}]{Yu2015}
Yu R, Russell C, Campbell NDF and Agapito L (2015) Direct, dense, and
  deformable: Template-based non-rigid 3d reconstruction from rgb video.
\newblock IEEE.
\newblock ISBN 978-1-4673-8391-2, pp. 918--926.
\newblock \doi{10.1109/ICCV.2015.111}.
\newblock \urlprefix\url{http://ieeexplore.ieee.org/document/7410468/}.

\bibitem[{Zanlungo et~al.(2011)Zanlungo, Ikeda and Kanda}]{zanlungo2011social}
Zanlungo F, Ikeda T and Kanda T (2011) Social force model with explicit
  collision prediction.
\newblock \emph{EPL (Europhysics Letters)} 93(6): 68005.

\bibitem[{Zhang and Parker(2011)}]{Zhang2011}
Zhang H and Parker LE (2011) 4-dimensional local spatio-temporal features for
  human activity recognition.
\newblock In: \emph{IEEE/RSJ International Conference on Intelligent Robots and
  Systems}. IEEE.
\newblock ISBN 978-1-61284-456-5, pp. 2044--2049.
\newblock \doi{10.1109/IROS.2011.6094489}.

\bibitem[{Zhang and Nakamura(2020)}]{Zhang2020}
Zhang T and Nakamura Y (2020) {PoseFusion: Dense RGB-D SLAM in Dynamic Human
  Environments}.
\newblock In: \emph{Springer Proceedings in Advanced Robotics}, volume~11. pp.
  772--780.
\newblock \doi{10.1007/978-3-030-33950-0_66}.

\bibitem[{Zhang et~al.(2019)Zhang, Zhang and Tang}]{Zhang2019}
Zhang Z, Zhang J and Tang Q (2019) {Mask R-CNN Based Semantic RGB-D SLAM for
  Dynamic Scenes}.
\newblock In: \emph{IEEE/ASME International Conference on Advanced Intelligent
  Mechatronics, AIM}, volume 2019-July.
\newblock ISBN 9781728124933, pp. 1151--1156.
\newblock \doi{10.1109/AIM.2019.8868400}.

\bibitem[{Zhi et~al.(2021)Zhi, Ott and Ramos}]{zhi2021probabilistic}
Zhi W, Ott L and Ramos F (2021) Probabilistic trajectory prediction with
  structural constraints.
\newblock In: \emph{2021 IEEE/RSJ International Conference on Intelligent
  Robots and Systems (IROS)}. IEEE, pp. 9849--9856.

\bibitem[{Zhou et~al.(2019)Zhou, Gao, Wang, Liu and Shen}]{Zhou2019}
Zhou B, Gao F, Wang L, Liu C and Shen S (2019) Robust and efficient quadrotor
  trajectory generation for fast autonomous flight.
\newblock \emph{IEEE Robotics and Automation Letters} 4: 3529--3536.
\newblock \doi{10.1109/LRA.2019.2927938}.

\bibitem[{Ziebart et~al.(2008)Ziebart, Maas, Bagnell, Dey
  et~al.}]{ziebart2008maximum}
Ziebart BD, Maas AL, Bagnell JA, Dey AK et~al. (2008) Maximum entropy inverse
  reinforcement learning.
\newblock In: \emph{Aaai}, volume~8. Chicago, IL, USA, pp. 1433--1438.

\bibitem[{Ziebart et~al.(2009)Ziebart, Ratliff, Gallagher, Mertz, Peterson,
  Bagnell, Hebert, Dey and Srinivasa}]{ziebart2009planning}
Ziebart BD, Ratliff N, Gallagher G, Mertz C, Peterson K, Bagnell JA, Hebert M,
  Dey AK and Srinivasa S (2009) Planning-based prediction for pedestrians.
\newblock In: \emph{IEEE/RSJ International Conference on Intelligent Robots and
  Systems}. IEEE, pp. 3931--3936.

\end{thebibliography}

\end{document}